\UseRawInputEncoding
\documentclass[a4paper,fleqn]{cas-sc}

\usepackage[utf8]{inputenc}

\usepackage[authoryear,longnamesfirst]{natbib}

\usepackage{amsmath}
\usepackage{amssymb}
\usepackage[utf8]{inputenc}

\usepackage{graphicx}
\usepackage{subcaption}

\usepackage{booktabs}   
\usepackage{tabularx}   
\usepackage{pdflscape}  
\usepackage{threeparttable} 

\usepackage{tikz}
\usepackage{hyperref}
\usetikzlibrary{arrows.meta, positioning, shapes.multipart, calc, fit, backgrounds}

\usepackage{enumitem}

\usepackage{xcolor}     
\usepackage{hyperref}   

\def\tsc#1{\csdef{#1}{\textsc{\lowercase{#1}}\xspace}}
\tsc{WGM}
\tsc{QE}
\tsc{EP}
\tsc{PMS}
\tsc{BEC}
\tsc{DE}

\begin{document}
\let\WriteBookmarks\relax
\def\floatpagepagefraction{1}
\def\textpagefraction{.001}

\shorttitle{SAFN}

\shortauthors{Datta et~al.}

\title[mode=title]{A Sparse-Attention Deep Learning Model Integrating Heterogeneous Multimodal Features for Parkinson’s Disease Severity Profiling}

\author[1]{\small Dristi Datta}
\author[1,3]{\small Tanmoy Debnath\texorpdfstring{\thanks{Corresponding author: tdebnath@csu.edu.au}}{ (Corresponding author: tdebnath@csu.edu.au)}}  
\author[2]{\small Minh Chau}
\author[1]{\small Manoranjan Paul}
\author[4]{\small Gourab Adhikary}
\author[1]{\small Md Geaur Rahman}  

\affiliation[1]{%
  organization={School of Computing, Mathematics, and Engineering, Charles Sturt University},
  addressline={NSW},
  country={Australia}
}

\affiliation[2]{%
   organization={School of Dentistry and Medical Sciences, Charles Sturt University},
   addressline={NSW},
   country={Australia}
}

\affiliation[3]{%
   organization={AI and Cyber Futures Centre (AICF)},
   addressline={NSW},
   country={Australia}
}

\affiliation[4]{%
   organization={Hawkins Clinic General Medical Practice},
   addressline={South Australia},
   country={Australia}
}

\footnotetext[1]{Corresponding author: tdebnath@csu.edu.au}



\begin{abstract}
Characterising the heterogeneous presentation of Parkinson’s disease (PD) requires integrating biological and clinical markers into a unified predictive framework. While multimodal data offer complementary information, many existing computational models struggle with interpretability, class imbalance, or effective fusion of high-dimensional imaging with tabular clinical features. To address these limitations, we propose the Class-Weighted Sparse-Attention Fusion Network (SAFN), which is an interpretable deep learning framework for robust multimodal profiling. The proposed SAFN framework integrates MRI Cortical Thickness, MRI Volumetric measures, comprehensive clinical assessments, and demographic variables through modality-specific encoders and a symmetric cross-attention mechanism that captures non-linear interactions between imaging- and clinical-derived representations. A sparsity-constrained attention-gating fusion layer dynamically prioritises informative modalities, while a Class-Balanced Focal Loss ($\beta=0.999,\gamma=1.5$) mitigates dataset imbalance without synthetic oversampling. Evaluated on 703 participants (570 PD, 133 HC) from the Parkinson’s Progression Markers Initiative (PPMI) using subject-wise five-fold cross-validation, the SAFN achieves an accuracy of 0.98~$\pm$~0.02 and a PR--AUC of 1.00~$\pm$~0.00, outperforming established ML and DL baselines. The model’s interpretability analysis reveals a clinically coherent decision process: modality gating assigns roughly 60\% of predictive weight to clinical assessments, consistent with Movement Disorder Society diagnostic principles, while MRI-derived features provide complementary stratification. The SAFN thus delivers a reproducible, transparent multimodal modelling paradigm that bridges high-performance AI with clinical reasoning, offering a trustworthy foundation for computational profiling in neurodegenerative disease.
\end{abstract}



\maketitle

\section{Introduction}

Parkinson’s disease (PD) is a multifaceted, progressive neurodegenerative disorder that significantly diminishes quality of life. It is clinically characterised by motor symptoms such as bradykinesia, rigidity, tremor, and postural instability, alongside a wide range of non-motor manifestations including cognitive impairment, sleep disturbances, and autonomic dysfunction. Pathologically, PD is marked by the loss of dopaminergic neurons in the substantia nigra pars compacta (SNpc); however, overt motor symptoms typically emerge only after an estimated 50--70\% of these neurons have degenerated \citep{kempster2025understanding, twala2025ai}.

PD is also one of the fastest-growing neurological disorders worldwide. Contemporary epidemiological projections estimate that its global prevalence will exceed twenty-five million individuals by 2050, a 112\% increase from 2021, primarily driven by population ageing and sociodemographic transitions \citep{Li2025EOPD, Su2025}. Analyses from the Global Burden of Disease Study further illustrate sustained increases in incidence, mortality, and disability-adjusted life years, particularly among adults aged 55 years and older \citep{Peng2025}. Consequently, PD places a substantial and escalating burden on healthcare systems, patients, and caregivers.

Despite advances in therapeutic management, the comprehensive profiling of disease severity and phenotype remains challenging. Clinical evaluations, such as the Unified Parkinson’s Disease Rating Scale (MDS-UPDRS) and Hoehn \& Yahr staging, serve as the gold standard for functional assessment but rely heavily on symptomatic presentation. Fluctuations in patient state and overlap with atypical Parkinsonian syndromes further complicate diagnostic precision \citep{khan2024scoping, mattia2025deep, welton2024classification}. Moreover, while these scales quantify functional impairment, they do not directly reflect the underlying neuroanatomical degeneration. Heterogeneous disease trajectories and significant variability in clinical presentation additionally hinder the development of reliable, unified disease profiles.

These challenges underscore an urgent need for objective, scalable computational tools capable of synthesising heterogeneous data and integrating functional clinical signals with biological markers derived from neuroimaging. This has driven substantial interest in applying machine learning (ML) and deep learning (DL) methods to enhance the reproducibility and clinical relevance of PD characterisation. Recent work highlights the potential of automated biomarkers to detect neurodegenerative changes and relate them to clinical phenotypes. As a result, the field is increasingly converging on multimodal artificial intelligence (AI) frameworks designed to capture the complex, multisystem nature of PD.

To position our study within this evolving landscape, we review recent investigations spanning structural, functional, and microstructural MRI; radiomics; electrophysiological and EEG markers; behavioural and digital phenotyping; multimodal fusion strategies; differential diagnostic modelling; and systematic evaluations of computational PD biomarker research. Structural MRI research consistently demonstrates that PD involves distributed macrostructural alterations across the substantia nigra, basal ganglia, pallidum, thalamus, brainstem, cerebellum, and frontotemporal cortices \citep{Almgren2023, Alrawis2025, Basaia2024, Camacho2023, Camacho2024Exploiting, Hussain2025, Islam2024Machine, Li2024Parkinsons, Mahajan2025Deep, welton2024classification, Zhou2025Predictive}. Methodological approaches range from handcrafted morphometric feature sets to multi-centre CNNs whose saliency maps consistently prioritise deep grey matter and frontotemporal regions \citep{BalikciCicek2025Explainable, Camacho2024Exploiting}. More advanced architectures, including Swin Transformers and attention-based models, increasingly localise pathological cues to midbrain territories, though external generalisation remains a challenge \citep{Basaia2024, Hussain2025}. Complementary lines of work, including normative modelling, report atrophy patterns in subcortical, frontal, and cerebellar regions associated with disease presence \citep{Zheng2024Contrastive, Zhou2025Predictive}. However, a recurring insight is that models trained solely on MRI—whether voxelwise or feature-based—often suffer accuracy degradation on independent cohorts, reinforcing the need for frameworks that robustly integrate imaging with stable clinical information.

Beyond structural morphology, functional MRI (fMRI) and microstructural MRI provide complementary information. Resting-state connectivity analyses reveal alterations in sensorimotor and default mode networks \citep{Cao2020Radiomics, Mattia2025, Shi2022Machine}, while diffusion tensor imaging (DTI) and neuromelanin-sensitive MRI offer sensitive biomarkers of nigrostriatal degeneration \citep{Camacho2024Exploiting, Chen2024, Li2022Diffusion, Mattia2025, welton2024classification, Zhao2022Automated}. Electrophysiological studies using EEG additionally characterise PD as a neural oscillopathy, identifying abnormalities in alpha and beta rhythms \citep{Afonso2025, Bunterngchit2025, Jibon2024, Li2024, Zhao2024Interpretable}. Although these modalities yield rich biological insight, they are often explored in isolation or require complex preprocessing pipelines that limit clinical deployability.

Consequently, multimodal fusion has emerged as a leading strategy for integrated biomarker profiling. Frameworks combining structural MRI, genetic variants, and clinical assessments outperform unimodal baselines \citep{Dentamaro2024, Li2025EOPD, Mattia2025, Sar2025, Yang2025}. For instance, the PIDGN model fuses MRI with genetic SNPs using gated attention, illustrating the value of combining biological and clinical data \citep{Li2025EOPD}. Other approaches integrating voice, gait, sensor-derived, or tabular data similarly demonstrate strong predictive performance \citep{Colautti2025Systematic, Esan2025Association, Jin2025SHAP, Yang2025}. Despite these advances, several persistent limitations restrict clinical translation: (i) modality dominance, where high-dimensional MRI features overshadow highly informative clinical or demographic inputs; (ii) lack of interpretability, particularly in black-box fusion models that obscure the relative contribution of each modality; and (iii) class imbalance, which biases model learning in real-world datasets with disproportionate PD prevalence.

To address these challenges, we developed the Class-Weighted Sparse-Attention Fusion Network (SAFN), a multimodal architecture that explicitly targets key limitations in existing PD models: heterogeneous feature structures, modality dominance, and limited interpretability. SAFN integrates MRI Cortical Thickness, MRI Volumetric features, clinical assessments, and demographic variables using modality-specific encoders to preserve within-modality structure, symmetric cross-attention to model clinically meaningful interactions between imaging and clinical streams, and a sparsity-constrained attention-gating fusion layer to prevent high-dimensional MRI features from overshadowing informative clinical signals. By incorporating Class-Balanced Focal Loss directly into the optimisation objective, SAFN mitigates dataset imbalance without synthetic resampling. This framework provides a balanced, transparent, and clinically aligned approach to multimodal PD profiling, bridging the gap between high-performance AI and interpretable decision-making.

Building on the limitations identified in current multimodal research, this study makes several key contributions:

\begin{enumerate}
    \item \textbf{A unified multimodal profiling framework.}  
    We introduce the Class-Weighted Sparse-Attention Fusion Network (SAFN), which selectively fuses heterogeneous MRI-derived and clinical modalities to address modality heterogeneity and prevent high-dimensional imaging features from dominating sparse but informative clinical signals.

    \item \textbf{Built-in class imbalance handling.}  
    SAFN incorporates Class-Balanced Focal Loss with EMA-stabilised AdamW optimisation to explicitly counter the severe PD--HC imbalance common in real-world cohorts, eliminating the need for synthetic oversampling and preserving minority-class integrity.

    \item \textbf{Clinically aligned interpretability.}  
    Learned modality gates provide intrinsic transparency by quantifying modality-level contributions, revealing that SAFN assigns approximately 60\% predictive weight to clinical assessments—consistent with established diagnostic and staging principles in PD.

    \item \textbf{A reproducible benchmarking protocol.}  
    We establish a rigorous evaluation pipeline to ensure fair and reproducible comparison, using identical preprocessing, stratified 5-fold cross-validation, and systematic evaluation against classical machine-learning and deep-learning baselines.

    \item \textbf{Integration of four complementary biomarker domains.}  
    The framework unifies MRI Cortical Thickness, MRI Volumetric features, comprehensive clinical assessments, and demographic factors to capture the multisystem and heterogeneous nature of PD within a single end-to-end predictive model.

    \item \textbf{Superior performance and architectural validation.}  
    Through extensive ablation studies and stratified evaluation, we demonstrate that SAFN’s architectural components—cross-attention, sparse gating, and class-balanced optimisation—collectively yield consistent improvements over strong ensemble and deep-learning baselines across accuracy, balanced accuracy, and PR-AUC metrics.
\end{enumerate}

The remainder of this paper is organised as follows. Section~\ref{sec:dataprep} describes the dataset, preprocessing procedures, and modality organisation. Section~\ref{sec:methodology} presents the methodological framework, including baseline models and the proposed SAFN architecture. Section~\ref{sec:results} reports the quantitative findings, including comparisons between SAFN and existing ML/DL baselines. Section~\ref{sec:discussion} interprets these results in the context of existing PD literature, highlighting clinical relevance, ablation study outcomes, modality-level contributions, and outlining directions for future work. Finally, Section~\ref{sec:conclusion} summarises the key insights of this study.

\section{Data Preparation}
\label{sec:dataprep}

\subsection{Data Sourcing}
Data were sourced from the Parkinson’s Progression Markers Initiative (PPMI), a multicentre, longitudinal biomarker study that provides harmonised multimodal data for Parkinson’s disease (PD) research. Imaging-derived and tabular features were retrieved using PPMI participant identifiers and integrated into a unified subject-level database.

The dataset encompassed four primary feature domains:
\begin{itemize}
    \item \textbf{MRI Cortical Thickness:} regional cortical thickness measures extracted from T1-weighted MRI using FreeSurfer~v7.1.1;
    \item \textbf{MRI Volumetric Features:} regional and global volumetric indices including grey matter, white matter, cerebellar structures, and intracranial volume;
    \item \textbf{Clinical Assessments:} motor and non-motor scales such as MDS-UPDRS, MoCA, NMSS, disease duration, and medication state;
    \item \textbf{Demographic Attributes:} age, sex, handedness, and education.
\end{itemize}

All modalities were mapped to a single participant identifier schema to ensure reproducibility and alignment across imaging and clinical sources.

\subsection{Data Cleaning, Filtering, and Harmonisation}
The initial dataset consisted of 703 participants (570 PD, 133 healthy controls). Integrity checks were performed to identify duplicate entries, mismatched identifiers, and inconsistent variable encodings. Variables with more than 20\% missingness were excluded. Continuous features with sporadic missing values were imputed using median replacement, while categorical attributes (e.g., sex, handedness) were label-encoded.

Participants with incomplete MRI acquisitions or failed FreeSurfer reconstruction were excluded from analysis. Outliers in continuous variables were screened using $z$-scores and visual histogram inspection to ensure consistent scaling across heterogeneous feature distributions. Given the inherent class imbalance in the cohort (approximately 81\% PD), no resampling or rebalancing was applied at the data level; instead, imbalance was addressed explicitly during model optimisation.

Following cleaning, features were grouped into the four modality-specific domains described above to support modality-aware modelling. All numerical variables were standardised using $z$-score normalisation (zero mean, unit variance). MRI harmonisation across scanner sites followed PPMI’s standardised acquisition protocols, as verified against the PPMI MRI Technical Operations Manual, ensuring consistency across multi-centre data.

\subsection{Finalised Files for Analysis}
The final dataset used for modelling comprised:
\begin{itemize}
    \item 70 MRI cortical thickness features,
    \item 13 MRI volumetric indices,
    \item 409 clinical variables,
    \item 7 demographic attributes.
\end{itemize}

All features were consolidated into a unified tabular (\texttt{.csv}) format, with corresponding NIfTI files retained for traceability and cross-reference via participant identifier (\texttt{PATNO}). This standardised structure ensured consistent downstream preprocessing and fair comparison across all machine-learning and deep-learning models evaluated in this study.

\section{Methodology}
\label{sec:methodology}

The methodological framework was designed to develop a robust and interpretable PD classification system using multimodal biomedical data. This section is organised into five components: (i) an overall workflow describing the end-to-end processing and evaluation pipeline, (ii) model development and optimisation for the baseline ML and DL classifiers, (iii) a detailed description of the proposed Class-Weighted SAFN, (iv) model inference and evaluation using complementary performance metrics, and (v) a summary of the methodology. All models were trained using the same preprocessed feature set and stratified 5-fold cross-validation splits to ensure methodological consistency and enable fair comparison with the proposed SAFN model.

\subsection{Overall Workflow}
\begin{figure}
\centering
\includegraphics[width=\linewidth]{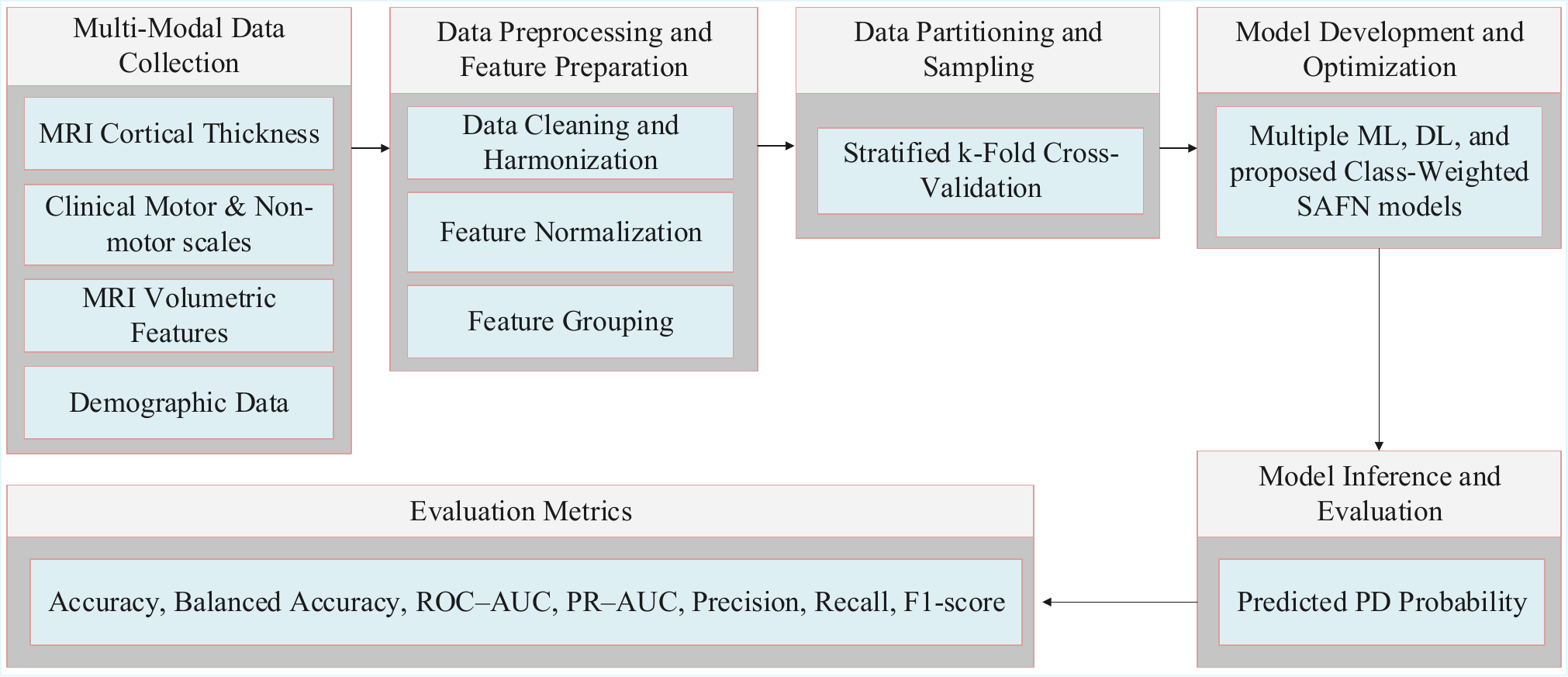}
\caption{Overall methodological workflow for Parkinson’s disease classification. 
Multimodal data undergo preprocessing, normalisation, and stratified sampling before model training. 
Machine-learning, deep-learning, and the proposed Class-Weighted SAFN models are trained under identical experimental settings and evaluated using Accuracy, Balanced Accuracy, ROC--AUC, PR--AUC, Precision, Recall, and F1-score metrics.}
\label{fig:methodology}
\end{figure}

The end-to-end workflow is illustrated in Fig.~\ref{fig:methodology}. Multimodal inputs comprised MRI cortical thickness features, MRI volumetric indices, clinical motor and non-motor scales, and demographic variables, all harmonised under a unified subject identifier and merged into a single tabular dataset (see Section~\ref{sec:dataprep}).

All preprocessing steps required for model training (imputation, encoding, and standardisation) were completed during data preparation. Within each cross-validation fold, imputers and scalers were refitted exclusively on the training subset to prevent data leakage and subsequently applied to the validation partition. Features were organised into four modality-aware groups—MRI Cortical Thickness, MRI Volumetric, Clinical, and Demographic—ensuring consistent structured inputs for all baseline models and enabling modality-informed encoding within SAFN.

Stratified $k$-fold cross-validation ($k=5$) was used to ensure balanced evaluation, with each fold serving once as the validation set while the remaining folds formed the training set. For participants with multiple entries, \texttt{StratifiedGroupKFold} enforced subject-level independence. Deep-learning models, including SAFN, were trained with early stopping based on validation loss, and final results were reported as mean~$\pm$~SD across folds.

\subsection{Model Development and Optimization}
A range of baseline models was implemented to benchmark performance, with their full hyperparameter settings summarised in Table~\ref{tab:hyperparams}. 
The ML baselines included logistic regression, SVM, random forest, XGBoost, LightGBM, and KNN, while the DL baselines comprised a fully connected ANN and a 1D--CNN. 
All models were trained using the same preprocessed feature set and stratified cross-validation splits to ensure fair and controlled comparison. 
Hyperparameters were tuned through internal validation, and the DL models were optimised using the Adam or AdamW optimisers with early stopping. 
The proposed Class-Weighted SAFN integrates self-attention encoding, symmetric cross-attention, and sparsity-regularised fusion, and addresses class imbalance using class-weighted loss functions rather than oversampling.

\begin{table}[ht!]
\centering
\caption{Hyperparameter settings for baseline ML, DL, and the proposed SAFN models. 
Symbol definitions: $C$ = inverse regularisation strength; $\gamma$ = RBF kernel width; 
$\lambda,\alpha$ = L2/L1 regularisation terms; $p$ = Minkowski distance exponent; 
$D_{\text{model}}$ = embedding dimension.}
\label{tab:hyperparams}
\small
\setlength{\tabcolsep}{6pt}
\begin{tabular}{@{}p{3.4cm}p{9.3cm}@{}}
\toprule
\textbf{Model} & \textbf{Key Hyperparameters} \\ 
\midrule

Logistic Regression 
& L2 regularisation; $C=1.0$; Solver: \texttt{liblinear}; Class weight: balanced; 
Max iterations: 2000; Random seed: 42; Preprocessing: median imputation + standardisation; categorical one-hot encoding \\

\midrule
SVM 
& Kernel: RBF; $C=1.0$; $\gamma=\text{scale}$; Class weight: balanced; probability=True; 
Random seed: 42; same preprocessing as above \\

\midrule
Random Forest 
& Trees: 600; Max depth: None; Min samples split: 5; Min leaf: 2; 
Class weight: balanced \\

\midrule
XGBoost 
& Estimators: 400; Learning rate: 0.05; Max depth: 6; 
Subsample: 0.8; Colsample: 0.8; Regularisation: $\lambda=1$, $\alpha=0$ \\

\midrule
LightGBM 
& Estimators: 400; Learning rate: 0.05; Max depth: 6; Num leaves: 31; 
Subsample: 0.8; Colsample by tree: 0.8; Regularisation: $\lambda=1$, $\alpha=0$; 
Objective: binary; Class weight: balanced; Random seed: 42 \\

\midrule
KNN 
& Neighbours: 7; Weights: \texttt{distance}; Metric: Minkowski ($p=2$, Euclidean); 
Standardised numeric inputs; one-hot encoded categoricals \\

\midrule
ANN 
& Hidden layers: [128, 64]; Activation: ReLU; Dropout: 0.4; 
Loss: BCEWithLogits; Class balancing via \texttt{pos\_weight}; 
Optimizer: Adam ($\eta=10^{-3}$); Batch size: 64; Epochs: 50; 
Early stopping patience: 8 \\

\midrule
1D-CNN 
& Conv layers: [32, 64]; Kernel sizes: [5, 3]; Activation: ReLU; 
Global average pooling; Dense layer: 64; Dropout: 0.25; 
Optimizer: AdamW ($\eta=2\times10^{-4}$, weight decay=$10^{-4}$); 
Batch size: 64; Epochs: 60; Early stopping patience: 10; 
Class balancing via \texttt{pos\_weight} \\ 

\midrule
Proposed SAFN 
& $D_{\text{model}}=64$; Heads: 4; Layers: 2; Dropout: 0.3; 
Optimizer: AdamW ($\eta=2\times10^{-4}$, weight decay=$10^{-4}$); 
Scheduler: Warmup--Cosine; EMA=0.999; Epochs: 60; 
Early stopping patience: 12 \\

\bottomrule
\end{tabular}
\end{table}

\subsection{Proposed Class-Weighted Sparse-Attention Fusion Network}

\subsubsection{Overview}
The proposed Class-Weighted SAFN aims to classify PD using heterogeneous tabular modalities, including MRI Cortical Thickness features, MRI Volumetric features, clinical motor and non-motor scales, and demographic variables. Conventional early or late fusion strategies often (i) treat all modalities uniformly despite differences in predictive value and (ii) perform suboptimally under class imbalance between PD and HC subjects. SAFN mitigates these limitations through two key innovations: (i) an attention-based gated fusion module with sparsity regularization that selectively emphasizes informative modalities while suppressing noise; and (ii) an imbalance-aware objective based on class-balanced focal loss, which re-weights effective class contributions and focuses training on difficult samples. Additionally, a symmetric cross-attention mechanism is introduced between the Clinical and MRI Cortical Thickness streams to capture inter-modality dependencies often neglected by simple concatenation.

\subsubsection{Architecture Design}
\begin{figure}
    \centering
    \includegraphics[width=\linewidth]{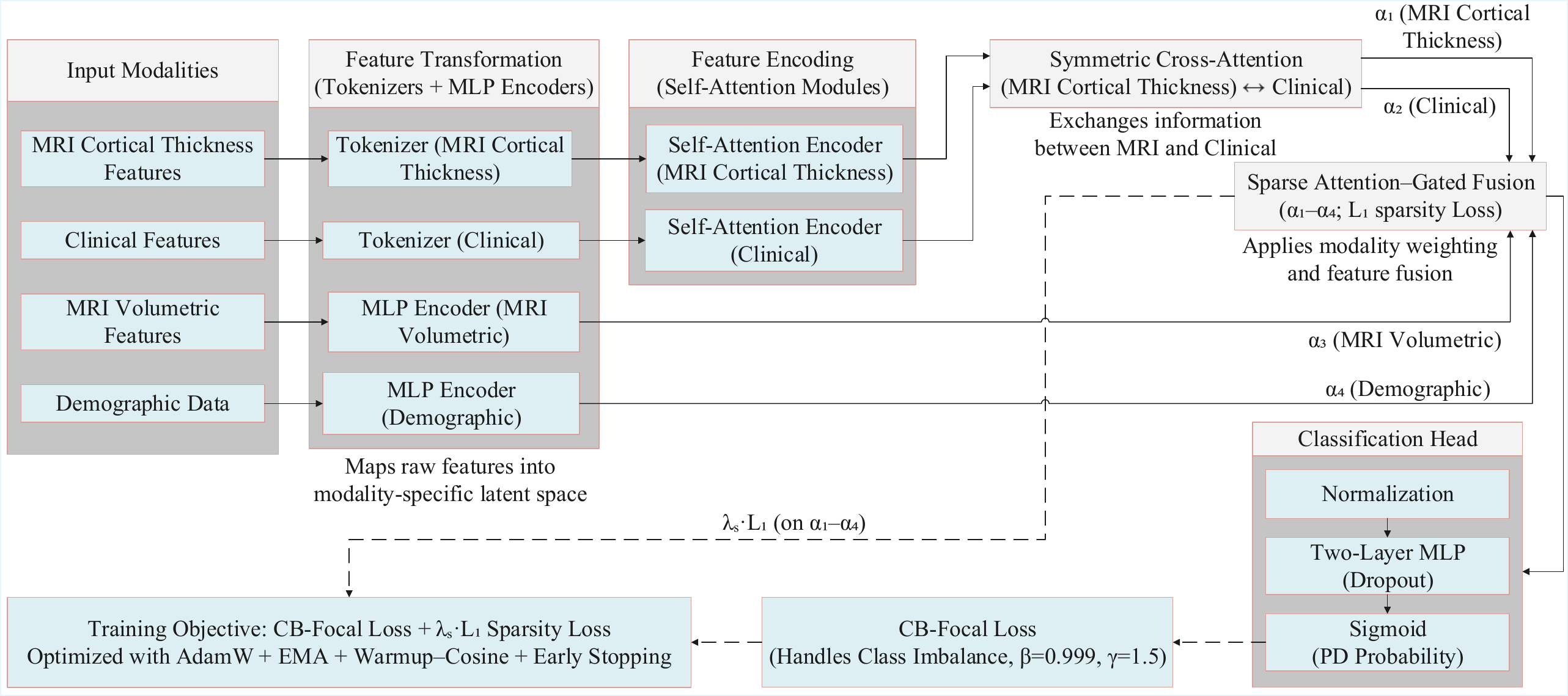}
    \caption{Architecture of the proposed Class-Weighted Sparse-Attention Fusion Network for Parkinson’s disease classification. Four input modalities—MRI Cortical Thickness, Clinical, MRI Volumetric, and Demographic features—are processed through modality-specific tokenizers or MLP encoders. MRI Cortical Thickness and Clinical embeddings interact via symmetric cross-attention, followed by sparse attention–gated multimodal fusion with learnable modality weights ($\alpha_1$–$\alpha_4$). The resulting fused representation $\mathbf{H}$ is passed to a classification head to predict PD probability. Solid arrows indicate data flow, while dashed arrows denote loss-related optimisation pathways.}

    \label{fig:SAFN_block}
\end{figure}

Fig.~\ref{fig:SAFN_block} illustrates the network structure, which comprises modality-specific tokenizers/encoders, a cross-attention module for inter-modality information exchange, a sparse attention–gated fusion block, and a normalized classification head.

\paragraph{(a) Modality-specific tokenization and encoders.}
Let $\mathbf{x}^{mri\_ct}\!\in\!\mathbb{R}^{F_{mri\_ct}}$, 
$\mathbf{x}^{mri\_vol}\!\in\!\mathbb{R}^{F_{mri\_vol}}$, 
$\mathbf{x}^{clin}\!\in\!\mathbb{R}^{F_{clin}}$, and 
$\mathbf{x}^{demo}\!\in\!\mathbb{R}^{F_{demo}}$ 
denote the feature vectors corresponding to the four modalities—MRI Cortical Thickness, MRI Volumetric, Clinical, and Demographic—after median imputation, standardisation, and categorical encoding. 
Here, $F_{mri\_ct}$, $F_{mri\_vol}$, $F_{clin}$, and $F_{demo}$ denote the number of features in each modality.

For the \emph{MRI Cortical Thickness} and \emph{Clinical} streams, each scalar feature is first mapped to a learnable embedding vector via a lightweight tokenizer, enabling feature-wise contextual modelling. 
This produces token sequences
$\mathbf{T}^{mri\_ct}\!\in\!\mathbb{R}^{F_{mri\_ct}\times D}$ and  
$\mathbf{T}^{clin}\!\in\!\mathbb{R}^{F_{clin}\times D}$,  
where $D$ is the shared embedding dimension across all modalities.  
Each token sequence is then processed by a two-layer Transformer encoder (four attention heads, GELU activation, dropout $p{=}0.4$), yielding contextualised representations:
\begin{align}
\mathbf{E}^{mri\_ct} &= \text{Enc}(\mathbf{T}^{mri\_ct}) \in \mathbb{R}^{F_{mri\_ct}\times D}, \\
\mathbf{E}^{clin}    &= \text{Enc}(\mathbf{T}^{clin})    \in \mathbb{R}^{F_{clin}\times D},
\end{align}
where $\text{Enc}(\cdot)$ denotes the modality-specific Transformer encoder. 
These encoded sequences subsequently undergo symmetric cross-attention and attention pooling (see part~(b)) to obtain fixed-length modality representations in $\mathbb{R}^D$.

For the \emph{MRI Volumetric} and \emph{Demographic} inputs, which are tabular and lower-dimensional, two-layer multilayer perceptron (MLP) encoders are employed to project the raw feature vectors directly into the same $D$-dimensional latent space:
\begin{equation}
\mathbf{Z}^{mri\_vol}=\text{MLP}(\mathbf{x}^{mri\_vol})\in\mathbb{R}^{D}, 
\qquad
\mathbf{Z}^{demo}=\text{MLP}(\mathbf{x}^{demo})\in\mathbb{R}^{D}.
\end{equation}

\paragraph{(b) Cross-attention between MRI Cortical Thickness and Clinical.}
To model complementary relationships between MRI Cortical Thickness and Clinical features,
symmetric cross-attention is applied \emph{after} the modality-specific Transformer encoders and
\emph{before} attention pooling. Accordingly, cross-attention operates on the encoded token sequences
$\mathbf{E}^{mri\_ct} \in \mathbb{R}^{F_{mri\_ct}\times D}$ and
$\mathbf{E}^{clin} \in \mathbb{R}^{F_{clin}\times D}$ defined in part~(a).

Formally, the cross-attended token representations are computed as
\begin{align}
\tilde{\mathbf{E}}^{mri\_ct} &= \text{CrossAttn}\!\left(\mathbf{E}^{mri\_ct} \leftarrow \mathbf{E}^{clin}\right), \\
\tilde{\mathbf{E}}^{clin}   &= \text{CrossAttn}\!\left(\mathbf{E}^{clin}   \leftarrow \mathbf{E}^{mri\_ct}\right),
\end{align}
where the notation $\mathbf{A} \leftarrow \mathbf{B}$ denotes that tokens in $\mathbf{A}$ act as
\emph{queries} while tokens in $\mathbf{B}$ provide the \emph{keys} and \emph{values}.
Here, $\text{CrossAttn}(\cdot)$ represents a multi-head cross-attention block followed by residual
connections, layer normalisation, and a position-wise feed-forward sublayer.

This bidirectional design enables both modalities to exchange complementary information:
MRI Cortical Thickness features are enriched by clinical context, while Clinical features incorporate
MRI-derived structural cues. Following cross-attention, an attention pooling operation aggregates
each token sequence into a fixed-length $D$-dimensional vector, yielding refined modality
representations
$\tilde{\mathbf{Z}}^{mri\_ct}, \tilde{\mathbf{Z}}^{clin} \in \mathbb{R}^{D}$
for subsequent fusion.

\paragraph{(c) Sparse attention--gated fusion.}
The four modality embeddings are concatenated and modulated by learnable sigmoid gates to obtain a fused multimodal representation. 
Let $\tilde{\mathbf{Z}}^{mri\_ct}, \tilde{\mathbf{Z}}^{clin}, \mathbf{Z}^{mri\_vol}, \mathbf{Z}^{demo} \in \mathbb{R}^{D}$ 
denote the modality-specific latent vectors obtained from parts~(a) and~(b). 
These vectors are first concatenated as
\begin{equation}
\mathbf{Z}
=
[\tilde{\mathbf{Z}}^{mri\_ct};\,\tilde{\mathbf{Z}}^{clin};\,\mathbf{Z}^{mri\_vol};\,\mathbf{Z}^{demo}]
\in \mathbb{R}^{4D},
\end{equation}
where the semicolon ``;'' denotes concatenation along the feature dimension and $D$ is the embedding dimensionality of each modality.

A modality-level gating vector is then computed as
\begin{equation}
\boldsymbol{\alpha}
=
\sigma(\mathbf{W}_g \mathbf{Z} + \mathbf{b}_g)
\in (0,1)^4,
\end{equation}
where $\mathbf{W}_g \in \mathbb{R}^{4 \times 4D}$ and $\mathbf{b}_g \in \mathbb{R}^{4}$ are learnable parameters, and
$\sigma(\cdot)$ denotes the element-wise sigmoid activation.
The vector
$\boldsymbol{\alpha} = [\alpha_1, \alpha_2, \alpha_3, \alpha_4]^\top$
encodes the relative importance of MRI Cortical Thickness, Clinical, MRI Volumetric, and Demographic modalities, respectively.

The fused representation is obtained by applying these gates to each modality embedding prior to concatenation:
\begin{equation}
\mathbf{H}
=
[\alpha_1\,\tilde{\mathbf{Z}}^{mri\_ct};\,
 \alpha_2\,\tilde{\mathbf{Z}}^{clin};\,
 \alpha_3\,\mathbf{Z}^{mri\_vol};\,
 \alpha_4\,\mathbf{Z}^{demo}]
\in \mathbb{R}^{4D}.
\end{equation}
Thus, each modality vector is first scaled by its corresponding gate $\alpha_j$ and then concatenated to form the fused multimodal representation $\mathbf{H}$.

To encourage selective utilisation of modalities, an $\ell_1$ sparsity penalty is imposed on the gating coefficients:
\begin{equation}
\mathcal{L}_{\mathrm{sparse}}
=
\lambda_s \sum_{j=1}^{4} |\alpha_j|,
\end{equation}
where $\lambda_s > 0$ controls the sparsity strength and promotes emphasis on the most informative modalities while suppressing weaker contributors.

\paragraph{(d) Classification head.}
The fused multimodal representation $\mathbf{H} \in \mathbb{R}^{4D}$ is first normalised using layer normalisation and then passed through a two-layer multilayer perceptron (MLP) with GELU activation and dropout regularisation to produce a scalar logit $s \in \mathbb{R}$. 
The final Parkinson’s disease (PD) probability is obtained via a sigmoid activation:
\begin{align}
s &= \text{MLP}\!\big(\text{LayerNorm}(\mathbf{H})\big), \\
\hat{y} &= \sigma(s),
\end{align}
where $\text{LayerNorm}(\cdot)$ denotes layer normalisation, 
$\text{MLP}(\cdot)$ represents the classification head, 
$\sigma(\cdot)$ is the sigmoid function, and 
$\hat{y} \in (0,1)$ denotes the predicted probability of PD.

\subsubsection{Imbalance-aware Objective}

To mitigate the inherent class imbalance between PD and HC samples, SAFN employs a Class-Balanced Focal Loss (CB-Focal), which integrates effective-number weighting with focal modulation to emphasise hard and underrepresented examples. Given the output logit $s$ from the classification head and its corresponding probability $p=\sigma(s)$, the per-sample loss is defined as:
\begin{equation}
\mathcal{L}_{\mathrm{CB\mbox{-}Focal}}
= - \alpha_1 (1-p)^\gamma\, y\log(p+\epsilon)
   - \alpha_0\, p^\gamma\, (1-y)\log(1-p+\epsilon),
\label{eq:cbfocal}
\end{equation}
where $y\!\in\!\{0,1\}$ denotes the ground-truth label (PD = 1, HC = 0), $\gamma$ is the focusing parameter controlling the emphasis on difficult samples, and $\epsilon{=}10^{-7}$ is a small constant for numerical stability (implemented via probability clamping to $[\,\epsilon,\,1-\epsilon\,]$).

The class-balancing coefficients $\alpha_c$ (with $\alpha_1$ for PD and $\alpha_0$ for HC) are computed using the \emph{effective number of samples}~\citep{cui2019class}:
\begin{equation}
\alpha_c \;\propto\; \frac{1-\beta}{1-\beta^{\,n_c}},
\qquad \beta\in(0,1), \quad n_c=\text{number of samples in class }c.
\end{equation}
In our implementation, the effective-number weights are computed \emph{per mini-batch} based on the batch-specific class counts $n_c$. This dynamic formulation preserves the principle of class-balanced reweighting while ensuring stable optimisation under mini-batch training. The resulting class weights are inversely proportional to the effective numbers ($w_{\text{pos}}\!\propto\!1/\mathrm{EN}_{\text{pos}}$, $w_{\text{neg}}\!\propto\!1/\mathrm{EN}_{\text{neg}}$), without enforcing the constraint $\alpha_0+\alpha_1=1$.

The overall training objective combines the CB-Focal component with the sparsity regularisation induced by the modality-gating layer:
\begin{equation}
\mathcal{L}_{\mathrm{total}}
= \mathcal{L}_{\mathrm{CB\mbox{-}Focal}} 
+ \lambda_s\, \mathcal{L}_{\mathrm{sparse}},
\end{equation}
where $\lambda_s>0$ controls the strength of the $\ell_1$ sparsity constraint applied to the modality gates.

\subsubsection{Training and Optimization}
All components are trained end-to-end using AdamW (learning rate $2\times10^{-4}$, weight decay $10^{-4}$, $\beta_1{=}0.9$, $\beta_2{=}0.999$, $\epsilon{=}10^{-8}$). 
A linear warmup over the first $10\%$ of optimization steps is followed by a cosine decay schedule that anneals the learning rate to zero. 
An exponential moving average (EMA, decay $0.999$) of model parameters is maintained throughout training; all validation and early-stopping decisions use the EMA weights for improved stability. 
To further stabilise updates, gradients are clipped to an $\ell_2$-norm of $1.0$. 
Training proceeds for up to $60$ epochs with early stopping (patience $12$), monitored using a composite validation metric defined as the mean of AUROC, balanced accuracy, and F1-score computed at a fixed threshold of $0.5$. 
The mini-batch size is set to $64$. 
Unless otherwise specified, layers follow the default PyTorch initialisation scheme. 
All hyperparameters were selected \textit{a priori} from standard ranges and kept constant across the $5$ folds to ensure a fair and reproducible comparison between SAFN and the baseline models.

\subsubsection{Cross-validation and Preprocessing}
A 5-fold stratified cross-validation strategy is employed to ensure that the PD/HC class proportions are preserved across folds. 
Within each fold, imputers, scalers, and label encoders are fitted \emph{only} on the training split and then applied to the validation split to prevent data leakage. 
When multiple observations belong to the same subject, StratifiedGroupKFold is used with subject identifiers, ensuring subject-level independence between training and validation sets. 
Evaluation metrics are reported at both a fixed decision threshold of 0.5 and the fold-specific best-F1 threshold, the latter selected by sweeping $t\in[0.05,\,0.95]$ to identify the threshold that maximises the F1-score.

\subsubsection{Interpretability via Attention Gates}
SAFN provides two complementary levels of interpretability. 
First, \emph{token-level} attention maps within the MRI (cortical thickness and volumetric) and clinical encoders highlight the most influential features contributing to the prediction. 
Second, the \emph{modality-level} gating coefficients $\boldsymbol{\alpha}$ quantify the relative contribution of each modality to the fused representation and final output. 
Aggregating $\boldsymbol{\alpha}$ across subjects yields a global measure of modality importance, typically showing stronger contributions from Clinical features, followed by MRI Cortical Thickness and MRI Volumetric inputs, with Demographic variables offering supportive context. 
These hierarchical attention mechanisms enable transparent inspection of the model's decision process, enhancing scientific interpretability and clinical trustworthiness without relying on external post-hoc explanation methods.

\subsection{Model Inference and Evaluation}
After model training, inference was performed on the held-out validation subset of each fold to assess generalization performance under consistent experimental conditions. 
For all models, the weights from the best validation epoch were used to generate predictions on the corresponding validation split. 
For each subject $i$, a scalar logit $s_i$ was obtained and converted into a probability $\hat{y}_i = \sigma(s_i)$, representing the predicted likelihood of PD. 
A fixed decision threshold of 0.5 served as the primary operating point, while an additional adaptive threshold was determined per fold by selecting the value of $t \in [0.05, 0.95]$ that maximised the F1-score. 
This dual-threshold strategy provides complementary insights into classifier calibration, precision–recall trade-offs, and robustness across validation folds.

\paragraph{Evaluation metrics.}
Model performance was quantified using standard binary classification metrics that jointly evaluate overall accuracy, discrimination capability, and balance between PD and HC predictions. 
Let $\mathrm{TP}$, $\mathrm{TN}$, $\mathrm{FP}$, and $\mathrm{FN}$ denote true positives, true negatives, false positives, and false negatives, respectively. 
The metrics are defined as:
\begin{align}
\text{Accuracy} &= \frac{\mathrm{TP} + \mathrm{TN}}{\mathrm{TP} + \mathrm{TN} + \mathrm{FP} + \mathrm{FN}}, \\
\text{Precision} &= \frac{\mathrm{TP}}{\mathrm{TP} + \mathrm{FP}}, \qquad
\text{Recall (Sensitivity)} = \frac{\mathrm{TP}}{\mathrm{TP} + \mathrm{FN}}, \\
\text{F1-score} &= 2 \times \frac{\text{Precision} \times \text{Recall}}{\text{Precision} + \text{Recall}}, \\
\text{Balanced Accuracy} &= \frac{1}{2} \left(
\frac{\mathrm{TP}}{\mathrm{TP} + \mathrm{FN}} +
\frac{\mathrm{TN}}{\mathrm{TN} + \mathrm{FP}}
\right).
\end{align}

Threshold-independent discrimination was further assessed using the area under the receiver operating characteristic curve (ROC-AUC) and the area under the precision–recall curve (PR-AUC):
\begin{align}
\text{ROC-AUC} &= \int_0^1 \mathrm{TPR}(x)\, \mathrm{dFPR}(x), \\
\text{PR-AUC} &= \int_0^1 \text{Precision}(r)\, \mathrm{dRecall}(r),
\end{align}
where $\mathrm{TPR}$ and $\mathrm{FPR}$ denote the true- and false-positive rates across all thresholds. 
All metrics were computed independently for each cross-validation fold and summarised as mean~$\pm$~standard deviation, ensuring a statistically fair comparison across models.

\paragraph{Visualization and reporting.}
For each model, fold-wise confusion matrices, ROC curves, and PR curves were generated to visualise class separation and calibration behaviour. 
Mean ROC and PR curves were computed by interpolating each fold-level curve over a uniform grid and averaging across folds, producing smooth aggregated representations of cross-validation performance. 
To maintain clarity and avoid redundancy, only the averaged confusion matrix, ROC curve, and PR curve are reported in the Results section. 
Final performance values are presented as mean~$\pm$~standard deviation for all evaluation metrics, enabling comprehensive comparison across the baseline ML models, DL models, and the proposed Class-Weighted SAFN. 
All evaluations were implemented using \textsf{scikit-learn}~(v1.5) and \textsf{PyTorch}~(v2.1), ensuring full reproducibility.

\subsection{Summary of Methodology}
In summary, the proposed framework integrates multimodal feature preprocessing, baseline ML and DL modelling, and the development of an interpretable Class-Weighted SAFN within a unified experimental setup. 
All models were trained using identical preprocessing procedures and stratified five-fold cross-validation to ensure methodological fairness and reproducibility. 
Evaluation metrics were computed consistently across all models, and performance is reported as mean~$\pm$~SD across folds. 
This coherent design establishes a robust foundation for objectively comparing both the predictive performance and interpretability of the proposed SAFN against conventional ML and DL approaches.

\section{Results}
\label{sec:results}

\begin{table}[ht!]
\centering
\caption{Performance comparison of traditional ML and deep learning models (5-fold CV). Values are mean~$\pm$~SD, rounded to two decimals.}
\label{tab:model_performance}
\small
\setlength{\tabcolsep}{4pt}
\begin{tabular}{@{}p{3.0cm}p{1.6cm}p{1.8cm}p{1.6cm}p{1.8cm}p{1.7cm}p{1.7cm}p{1.7cm}@{}}
\toprule
\textbf{Model} & \textbf{Accuracy} & \textbf{Balanced Acc} & \textbf{ROC-AUC} & \textbf{PR-AUC} & \textbf{Precision} & \textbf{Recall} & \textbf{F1} \\
\midrule
Logistic Regression         
    & 0.94 $\pm$ 0.01 & 0.93 $\pm$ 0.03 & 0.98 $\pm$ 0.00 & 0.99 $\pm$ 0.00 & 0.97 $\pm$ 0.01 & 0.95 $\pm$ 0.02 & 0.96 $\pm$ 0.01 \\
SVM     
    & 0.94 $\pm$ 0.01 & 0.89 $\pm$ 0.03 & 0.97 $\pm$ 0.01 & 0.99 $\pm$ 0.00 & 0.95 $\pm$ 0.01 & 0.97 $\pm$ 0.01 & 0.96 $\pm$ 0.01 \\
Random Forest    
    & 0.95 $\pm$ 0.06 & 0.95 $\pm$ 0.03 & 0.95 $\pm$ 0.03 & 0.98 $\pm$ 0.01 & 0.99 $\pm$ 0.01 & 0.95 $\pm$ 0.07 & 0.97 $\pm$ 0.04 \\
XGBoost 
    & 0.96 $\pm$ 0.04 & 0.96 $\pm$ 0.02 & 0.97 $\pm$ 0.01 & 0.98 $\pm$ 0.01 & 0.99 $\pm$ 0.01 & 0.96 $\pm$ 0.05 & 0.97 $\pm$ 0.03 \\
LightGBM 
    & 0.96 $\pm$ 0.01 & 0.95 $\pm$ 0.01 & 0.98 $\pm$ 0.01 & 0.98 $\pm$ 0.01 & 0.98 $\pm$ 0.01 & 0.98 $\pm$ 0.01 & 0.98 $\pm$ 0.01 \\
KNN    
    & 0.88 $\pm$ 0.01 & 0.83 $\pm$ 0.04 & 0.90 $\pm$ 0.04 & 0.97 $\pm$ 0.01 & 0.94 $\pm$ 0.02 & 0.91 $\pm$ 0.01 & 0.92 $\pm$ 0.01 \\
ANN    
    & 0.93 $\pm$ 0.03 & 0.91 $\pm$ 0.03 & 0.98 $\pm$ 0.01 & 0.99 $\pm$ 0.01 & 0.97 $\pm$ 0.01 & 0.94 $\pm$ 0.05 & 0.95 $\pm$ 0.02 \\
1dCNN   
    & 0.58 $\pm$ 0.20 & 0.61 $\pm$ 0.07 & 0.70 $\pm$ 0.06 & 0.90 $\pm$ 0.03 & 0.70 $\pm$ 0.25 & 0.57 $\pm$ 0.30 & 0.62 $\pm$ 0.31 \\
\addlinespace[2pt]
\textbf{Class-Weighted SAFN (Proposed)}
    & \textbf{0.98 $\pm$ 0.02} 
    & \textbf{0.97 $\pm$ 0.05} 
    & \textbf{0.98 $\pm$ 0.02} 
    & \textbf{1.00 $\pm$ 0.00} 
    & \textbf{0.99 $\pm$ 0.03} 
    & \textbf{0.99 $\pm$ 0.01} 
    & \textbf{0.99 $\pm$ 0.01} \\

\bottomrule
\end{tabular}

\vspace{2pt}
\raggedright\footnotesize
\textit{Notes.} PR-AUC, Precision, Recall, and F1 are reported for the PD (positive) class. Random Forest and SAFN were trained without SMOTE; SAFN addresses class imbalance via a class-weighted loss. 
All values are reported as mean $\pm$ SD across 5-fold cross-validation; due to fold-wise variability, mean $\pm$ SD intervals may exceed the theoretical bounds of $[0,1]$, although all fold-level metrics lie within valid ranges.

\end{table}

\begin{figure}
    \centering
    \includegraphics[width=0.95\linewidth]{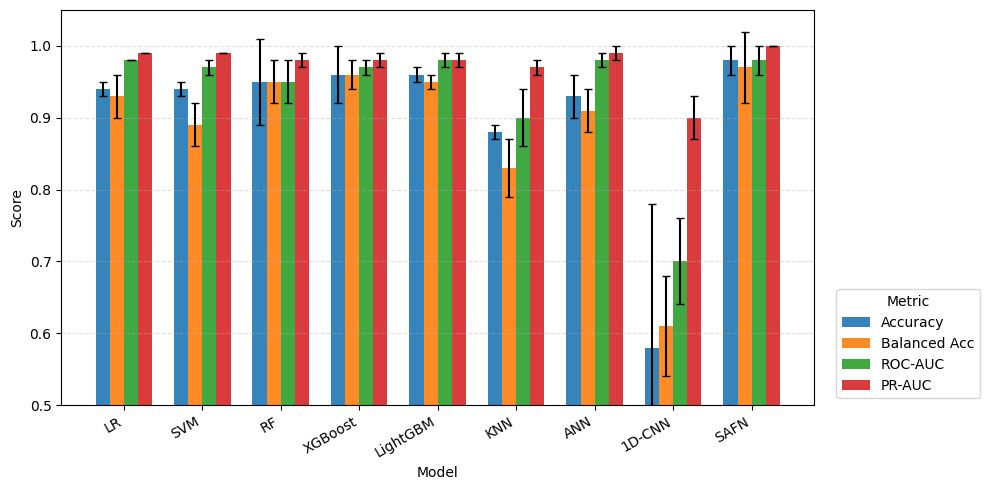}
    \caption{
        Grouped bar chart illustrating the performance of traditional machine-learning models and deep learning models across four key evaluation metrics: Accuracy, Balanced Accuracy, ROC-AUC, and PR-AUC. Error bars represent the standard deviation across five cross-validation folds. The proposed Class-Weighted SAFN exhibits consistently high performance with minimal variance compared to baseline models.
    }
    \label{fig:grouped_bar_performance}
\end{figure}

\begin{figure}
    \centering
    \includegraphics[width=0.7\linewidth]{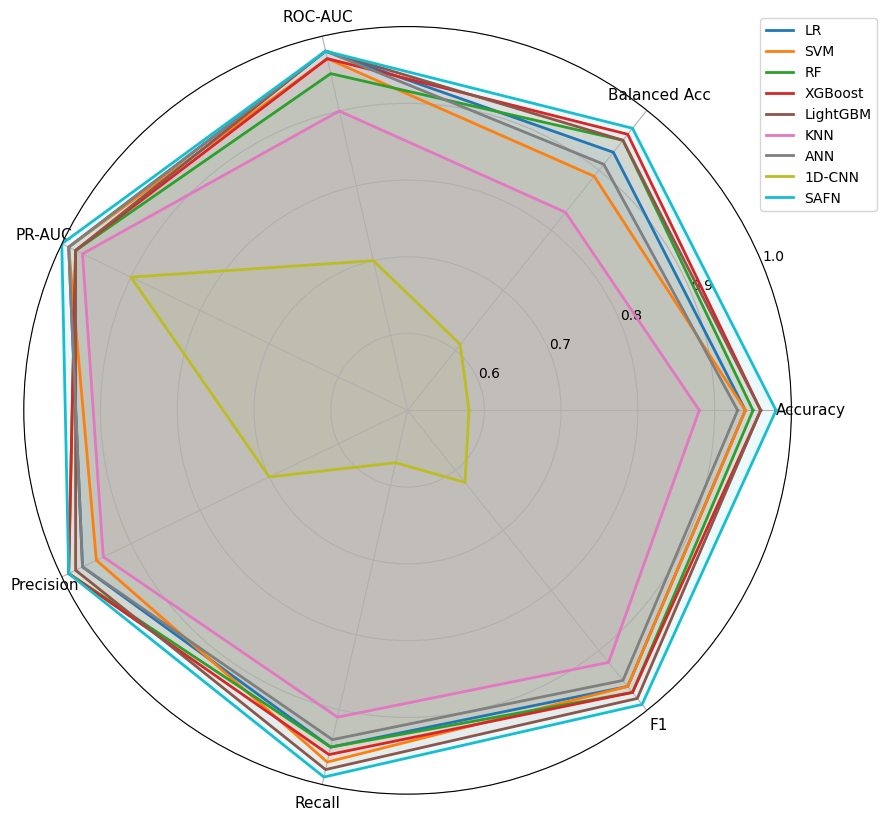}
    \caption{
        Radar chart comparing the performance of all baseline models and the proposed Class-Weighted SAFN across seven evaluation metrics (Accuracy, Balanced Accuracy, ROC-AUC, PR-AUC, Precision, Recall, and F1-score). Values correspond to mean scores over five-fold cross-validation. The SAFN model forms the outermost profile, indicating consistently superior performance.
    }
    \label{fig:radar_performance}
\end{figure}

\begin{figure*}
\centering
\begin{subfigure}[t]{0.32\textwidth}
  \centering
  \includegraphics[width=\linewidth]{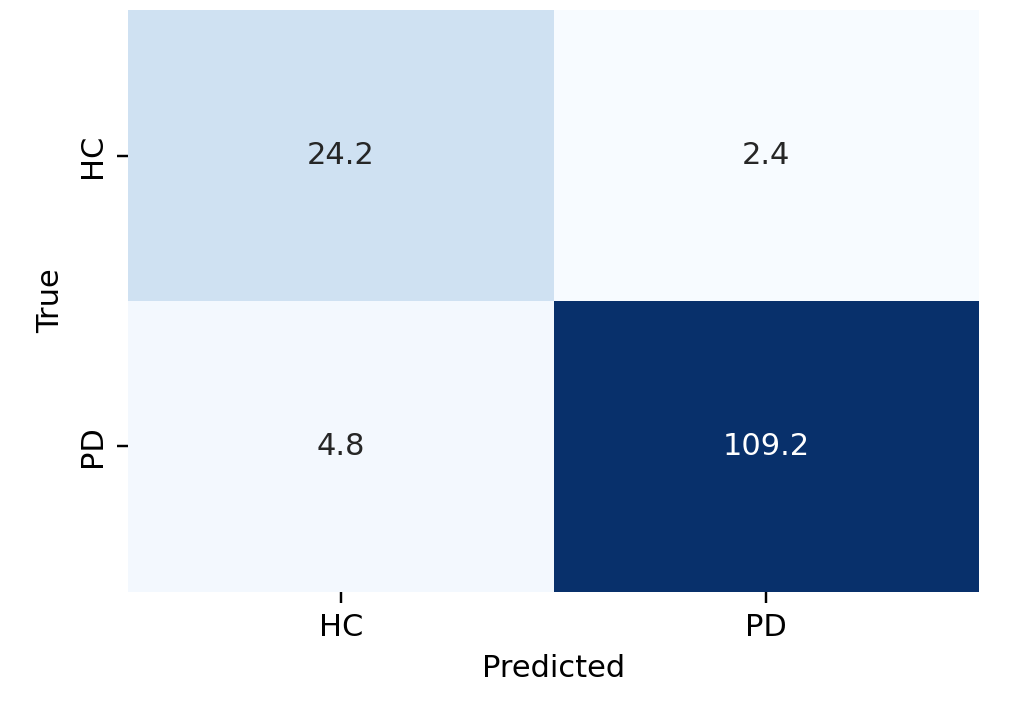}
  \caption{LogReg: Confusion matrix (avg)}
  \label{fig:lr_confmat}
\end{subfigure}
\begin{subfigure}[t]{0.32\textwidth}
  \centering
  \includegraphics[width=\linewidth]{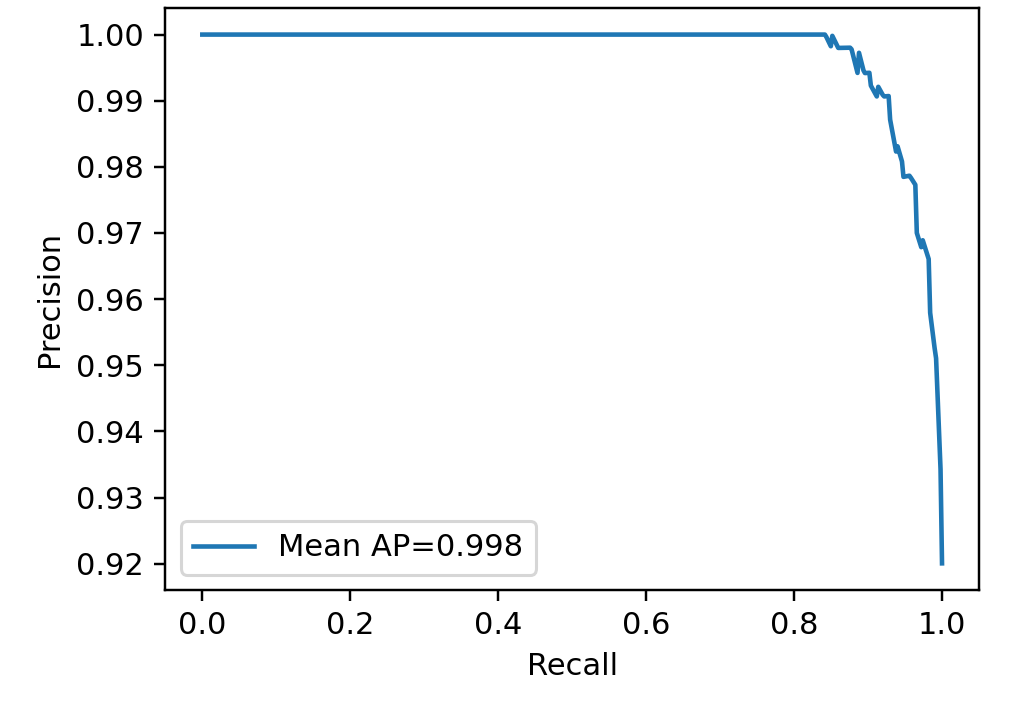}
  \caption{LogReg: PR curve (mean)}
  \label{fig:lr_pr}
\end{subfigure}
\begin{subfigure}[t]{0.32\textwidth}
  \centering
  \includegraphics[width=\linewidth]{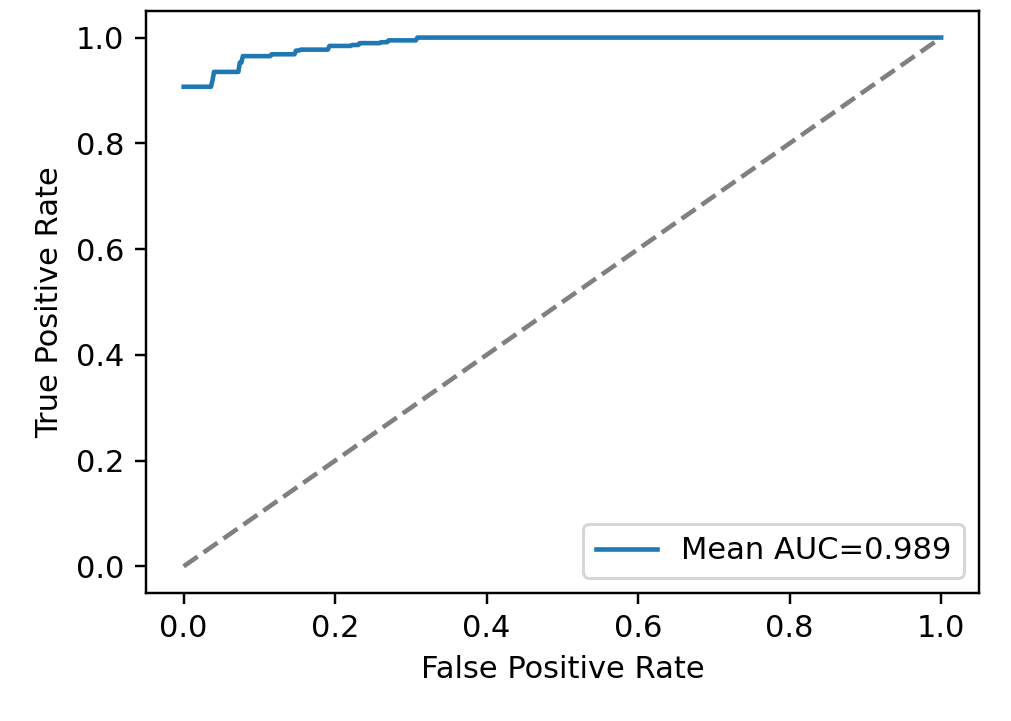}
  \caption{LogReg: ROC curve (mean)}
  \label{fig:lr_roc}
\end{subfigure}

\vspace{6pt} 

\begin{subfigure}[t]{0.32\textwidth}
  \centering
  \includegraphics[width=\linewidth]{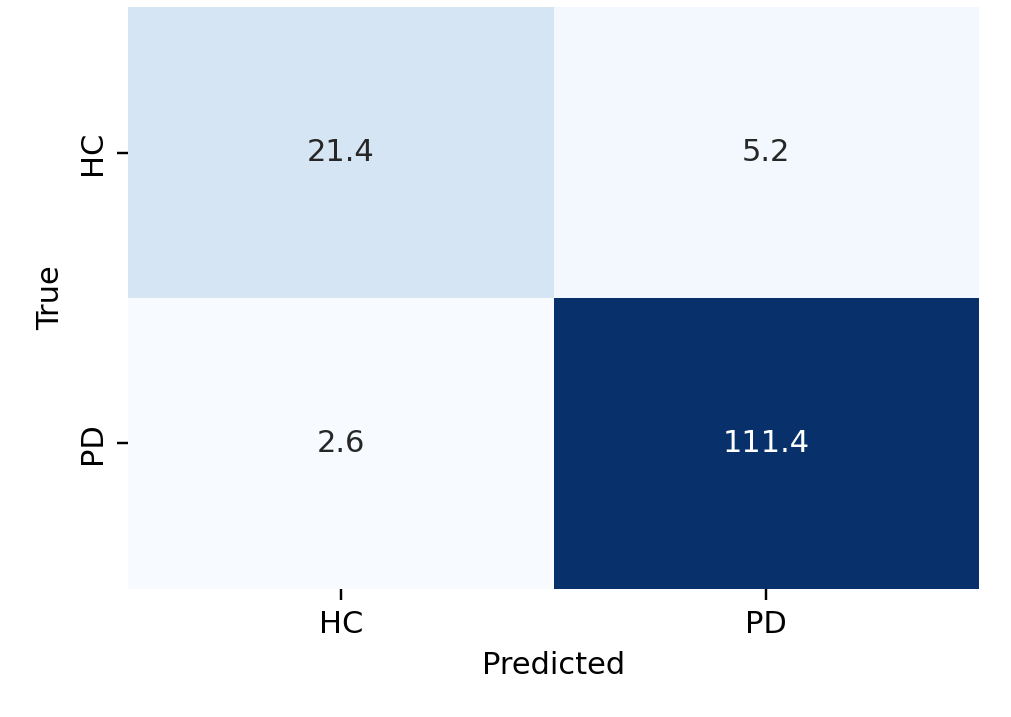}
  \caption{SVM: Confusion matrix (avg)}
  \label{fig:svm_confmat}
\end{subfigure}
\begin{subfigure}[t]{0.32\textwidth}
  \centering
  \includegraphics[width=\linewidth]{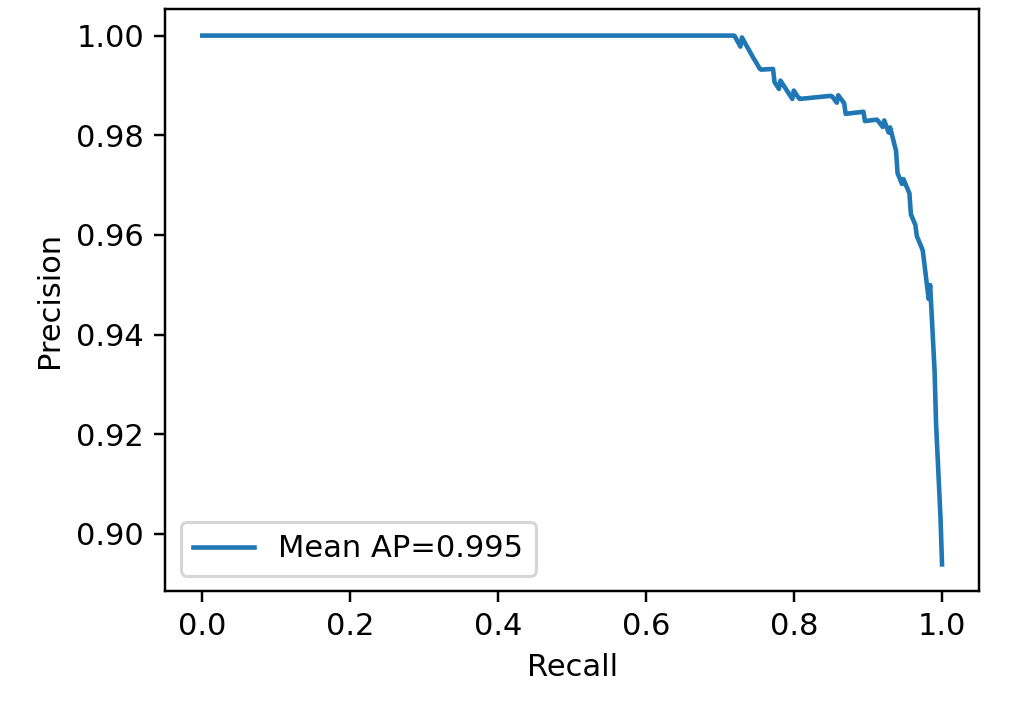}
  \caption{SVM: PR curve (mean)}
  \label{fig:svm_pr}
\end{subfigure}
\begin{subfigure}[t]{0.32\textwidth}
  \centering
  \includegraphics[width=\linewidth]{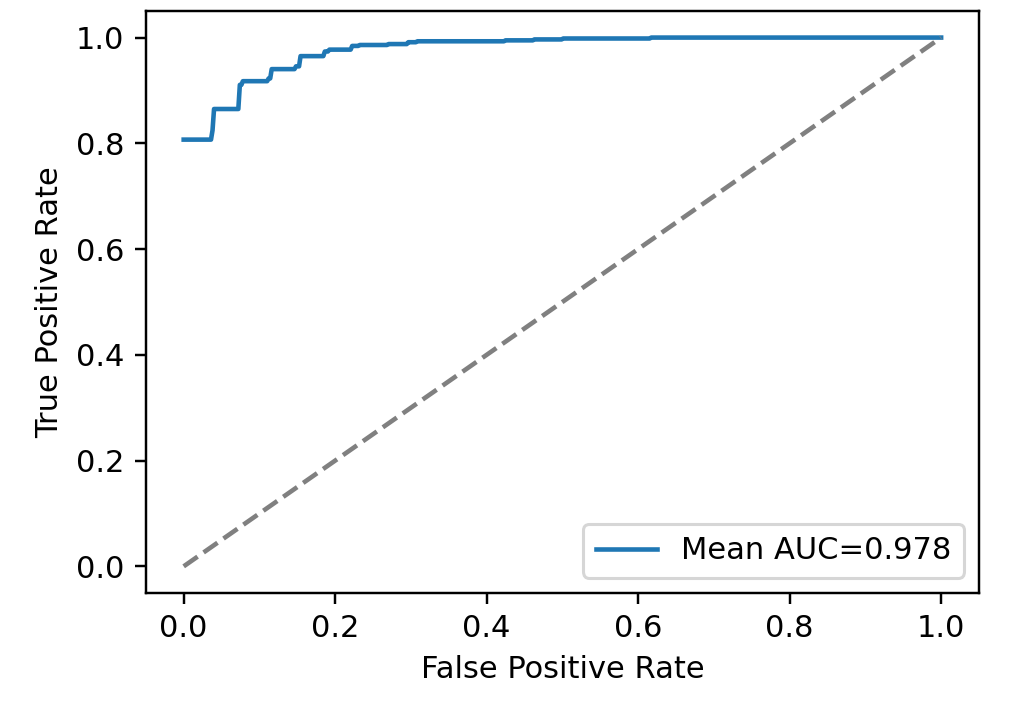}
  \caption{SVM: ROC curve (mean)}
  \label{fig:svm_roc}
\end{subfigure}

\vspace{6pt}

\begin{subfigure}[t]{0.32\textwidth}
  \centering
  \includegraphics[width=\linewidth]{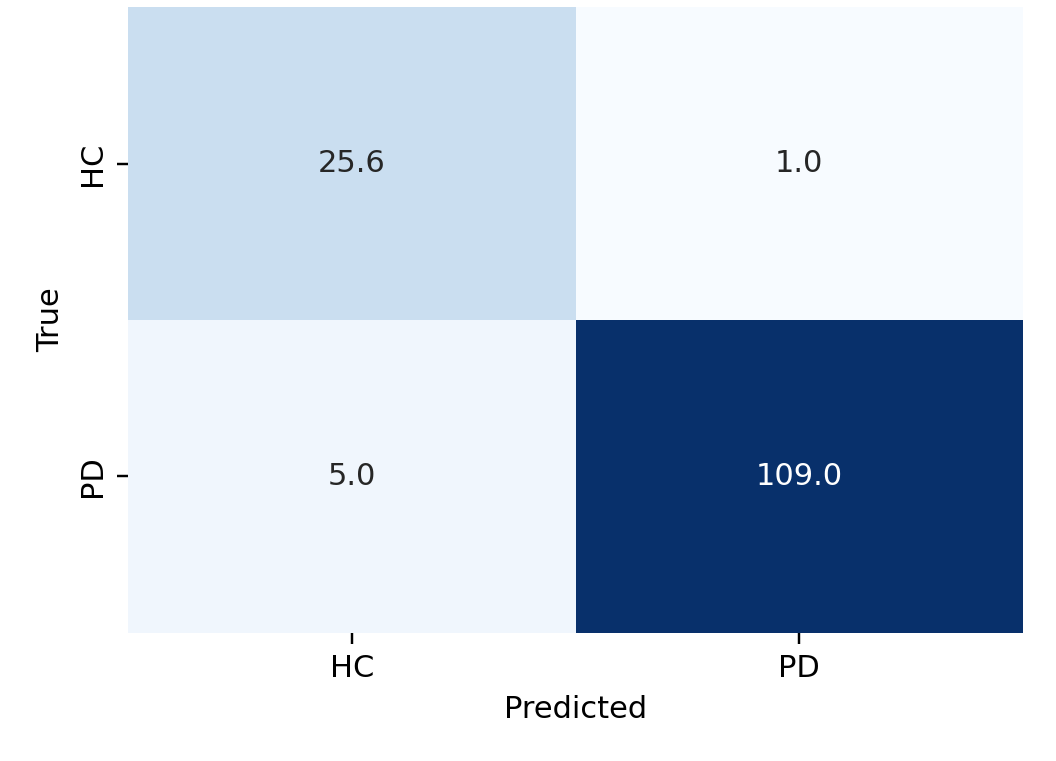}
  \caption{RF: Confusion matrix (avg)}
  \label{fig:rf_confmat}
\end{subfigure}
\begin{subfigure}[t]{0.32\textwidth}
  \centering
  \includegraphics[width=\linewidth]{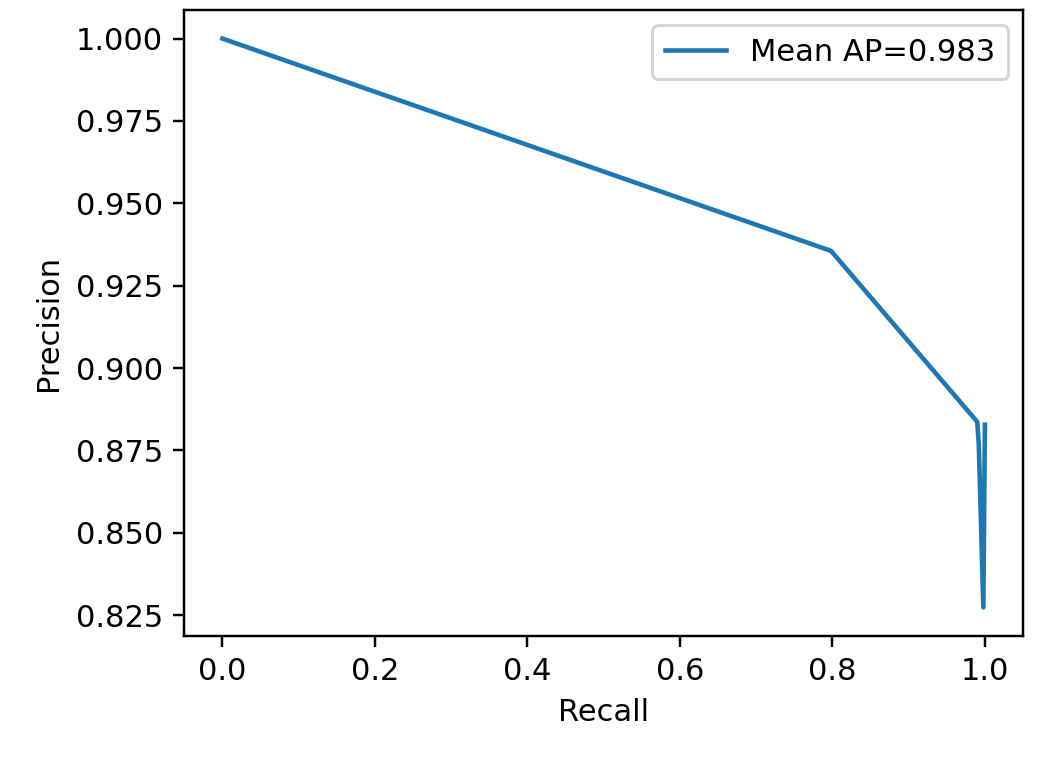}
  \caption{RF: PR curve (mean)}
  \label{fig:rf_pr}
\end{subfigure}
\begin{subfigure}[t]{0.32\textwidth}
  \centering
  \includegraphics[width=\linewidth]{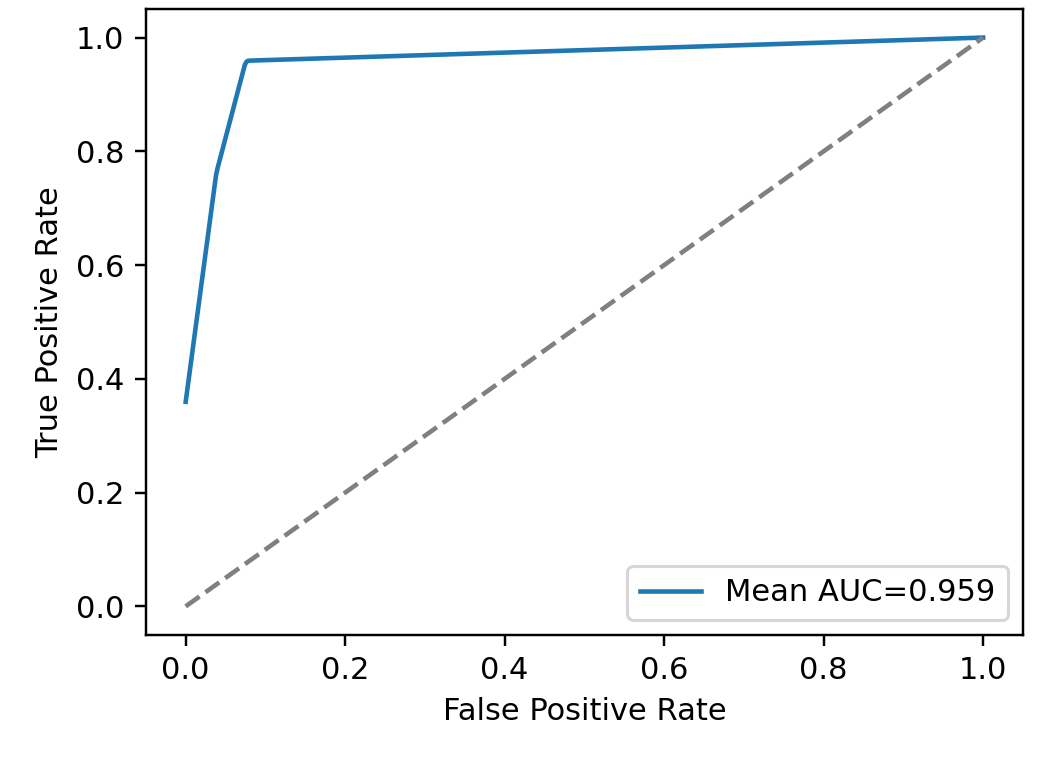}
  \caption{RF: ROC curve (mean)}
  \label{fig:rf_roc}
\end{subfigure}

\caption{Performance visualisation of traditional ML classifiers (Logistic Regression, SVM, and Random Forest). For each model, the figure displays the averaged confusion matrix, mean precision–recall curve, and mean ROC curve, computed across 5-fold cross-validation.}

\label{fig:ml_group1}
\end{figure*}

\begin{figure*}
\centering
\begin{subfigure}[t]{0.32\textwidth}
  \centering
  \includegraphics[width=\linewidth]{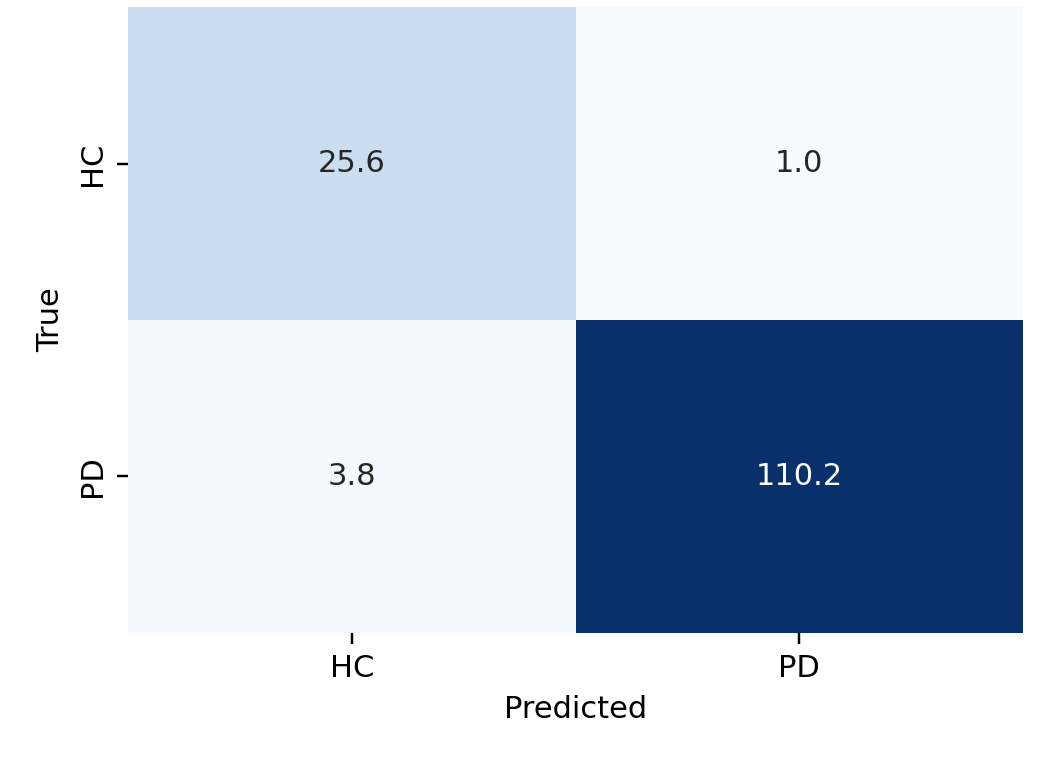}
  \caption{XGBoost: Confusion matrix (avg)}
  \label{fig:xgb_confmat}
\end{subfigure}
\begin{subfigure}[t]{0.32\textwidth}
  \centering
  \includegraphics[width=\linewidth]{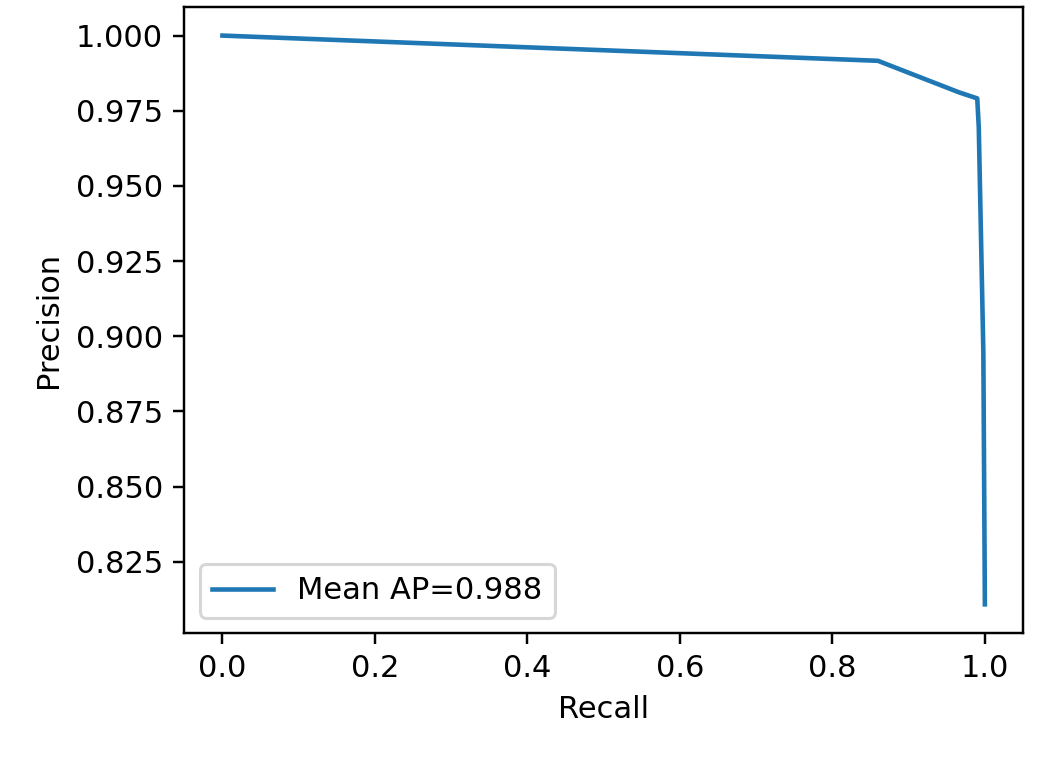}
  \caption{XGBoost: PR curve (mean)}
  \label{fig:xgb_pr}
\end{subfigure}
\begin{subfigure}[t]{0.32\textwidth}
  \centering
  \includegraphics[width=\linewidth]{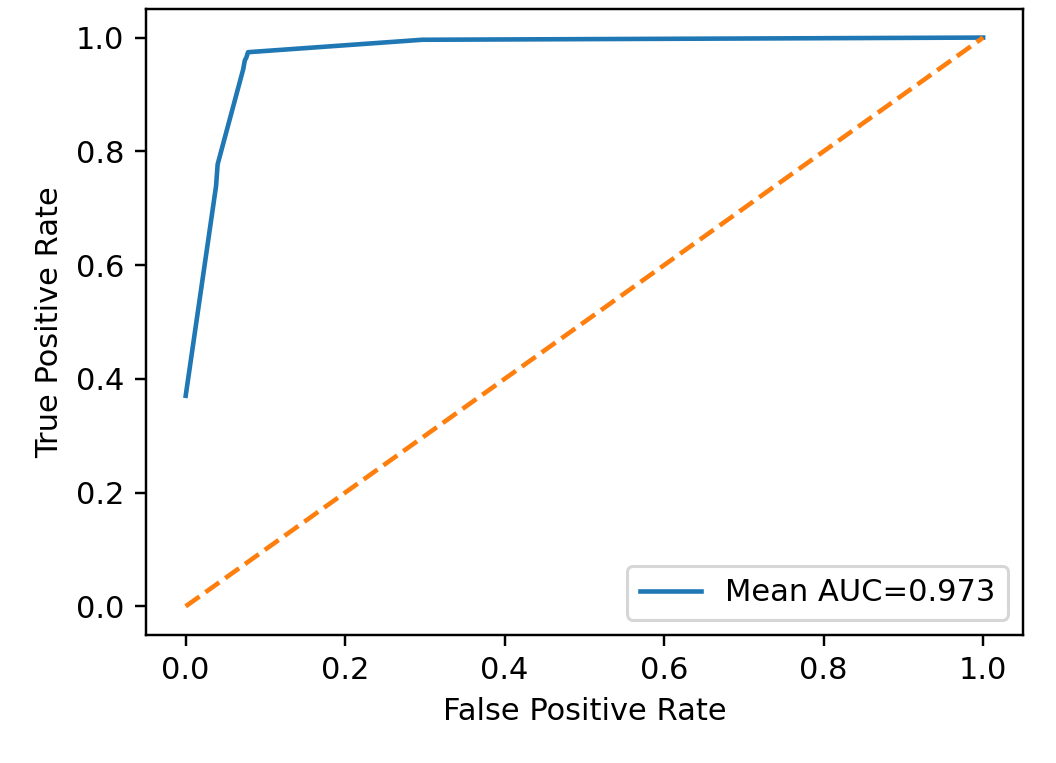}
  \caption{XGBoost: ROC curve (mean)}
  \label{fig:xgb_roc}
\end{subfigure}

\vspace{6pt} 

\begin{subfigure}[t]{0.32\textwidth}
  \centering
  \includegraphics[width=\linewidth]{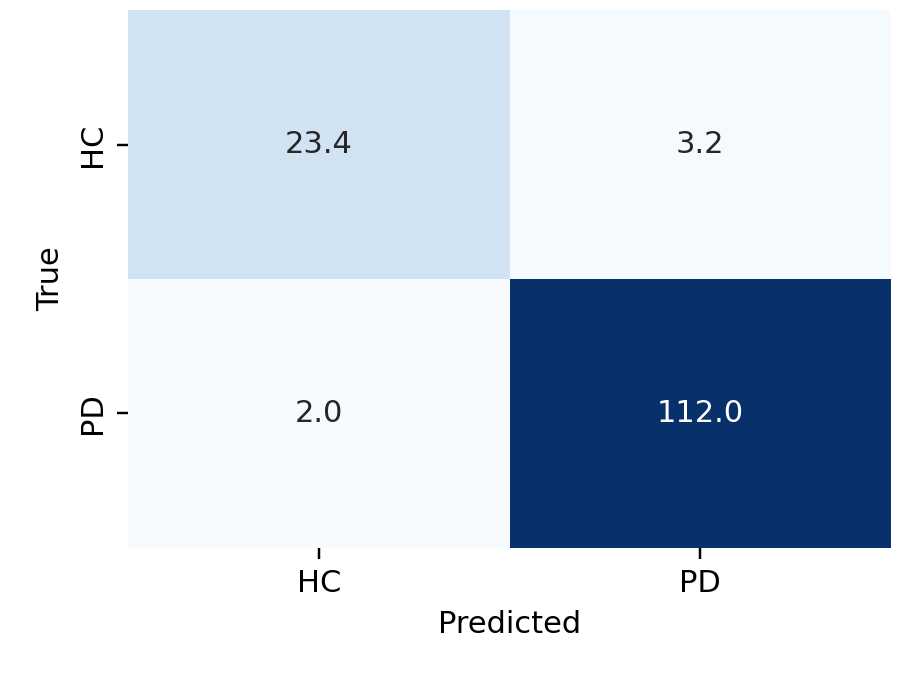}
  \caption{LightGBM: Confusion matrix (avg)}
  \label{fig:lgbm_confmat}
\end{subfigure}
\begin{subfigure}[t]{0.32\textwidth}
  \centering
  \includegraphics[width=\linewidth]{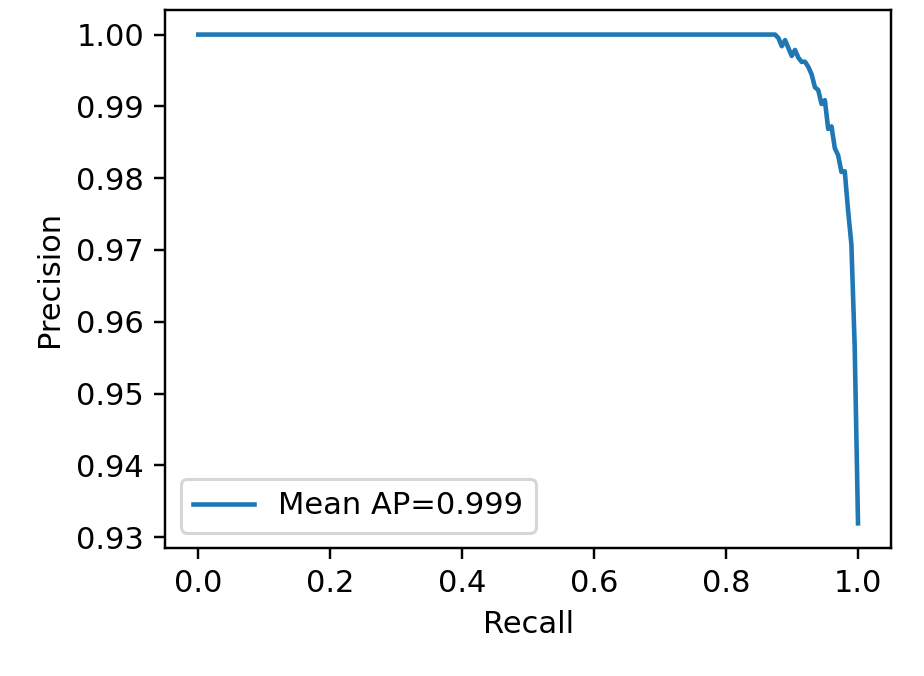}
  \caption{LightGBM: PR curve (mean)}
  \label{fig:lgbm_pr}
\end{subfigure}
\begin{subfigure}[t]{0.32\textwidth}
  \centering
  \includegraphics[width=\linewidth]{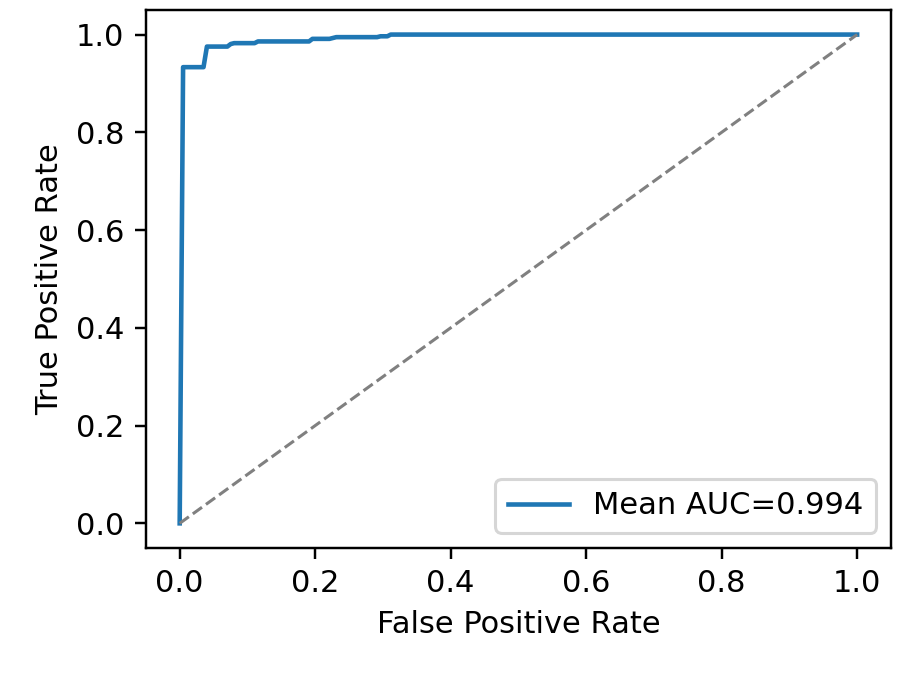}
  \caption{LightGBM: ROC curve (mean)}
  \label{fig:lgbm_roc}
\end{subfigure}

\vspace{6pt}

\begin{subfigure}[t]{0.32\textwidth}
  \centering
  \includegraphics[width=\linewidth]{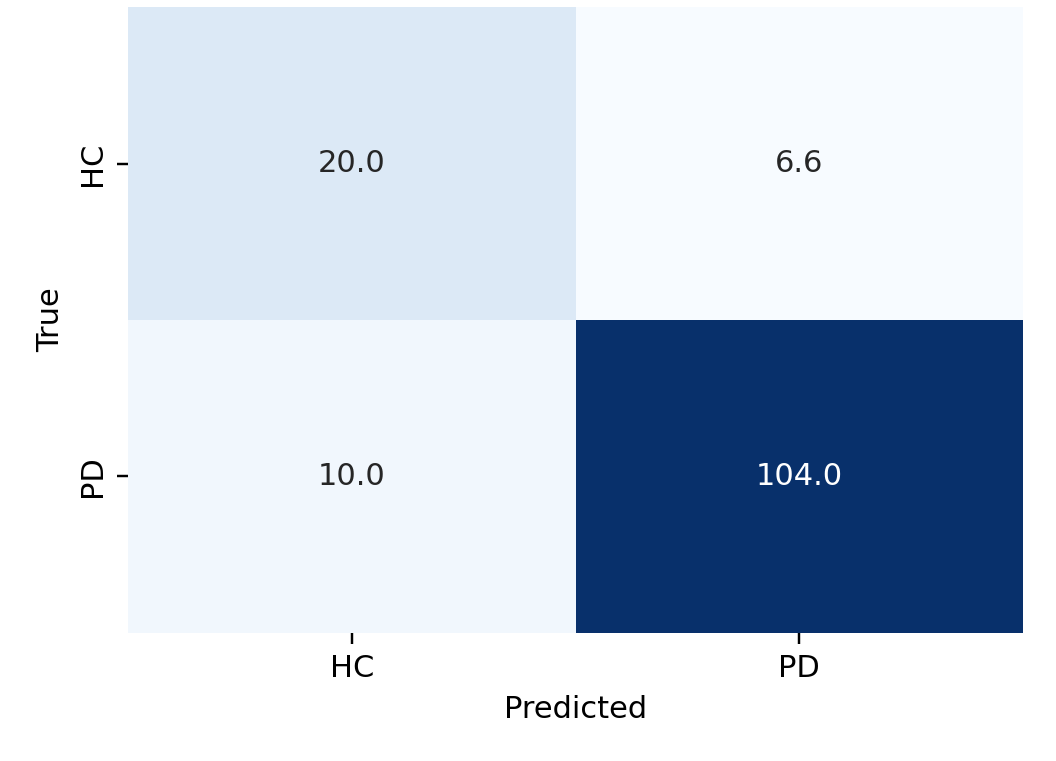}
  \caption{KNN: Confusion matrix (avg)}
  \label{fig:knn_confmat}
\end{subfigure}
\begin{subfigure}[t]{0.32\textwidth}
  \centering
  \includegraphics[width=\linewidth]{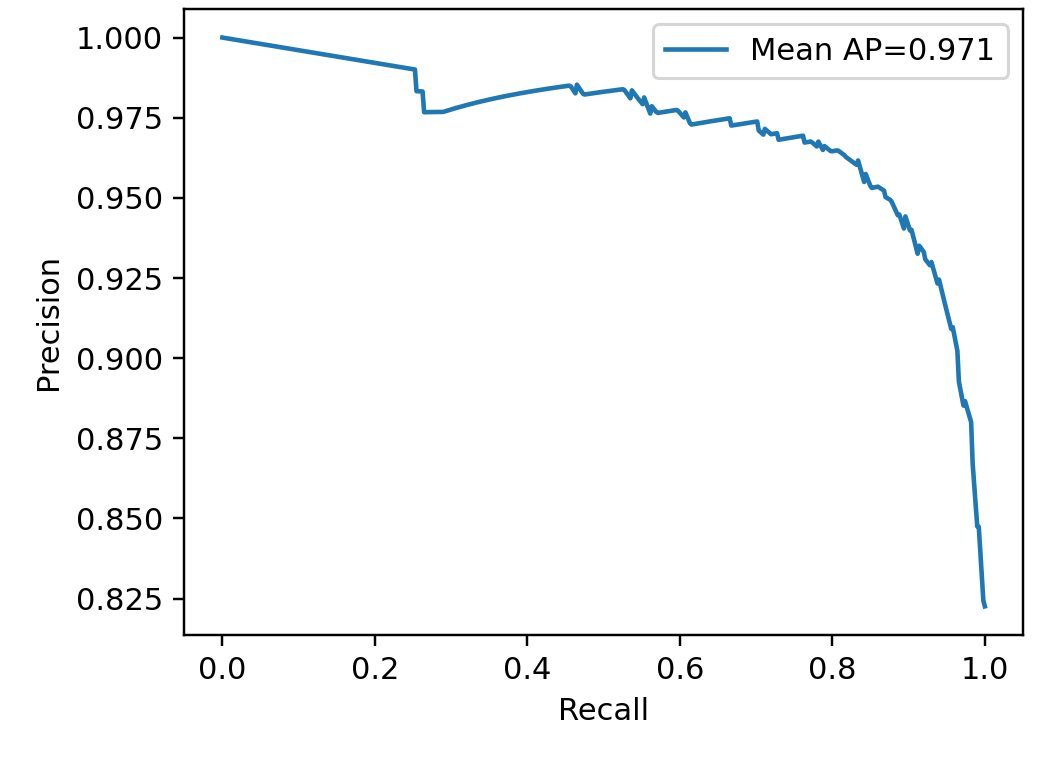}
  \caption{KNN: PR curve (mean)}
  \label{fig:knn_pr}
\end{subfigure}
\begin{subfigure}[t]{0.32\textwidth}
  \centering
  \includegraphics[width=\linewidth]{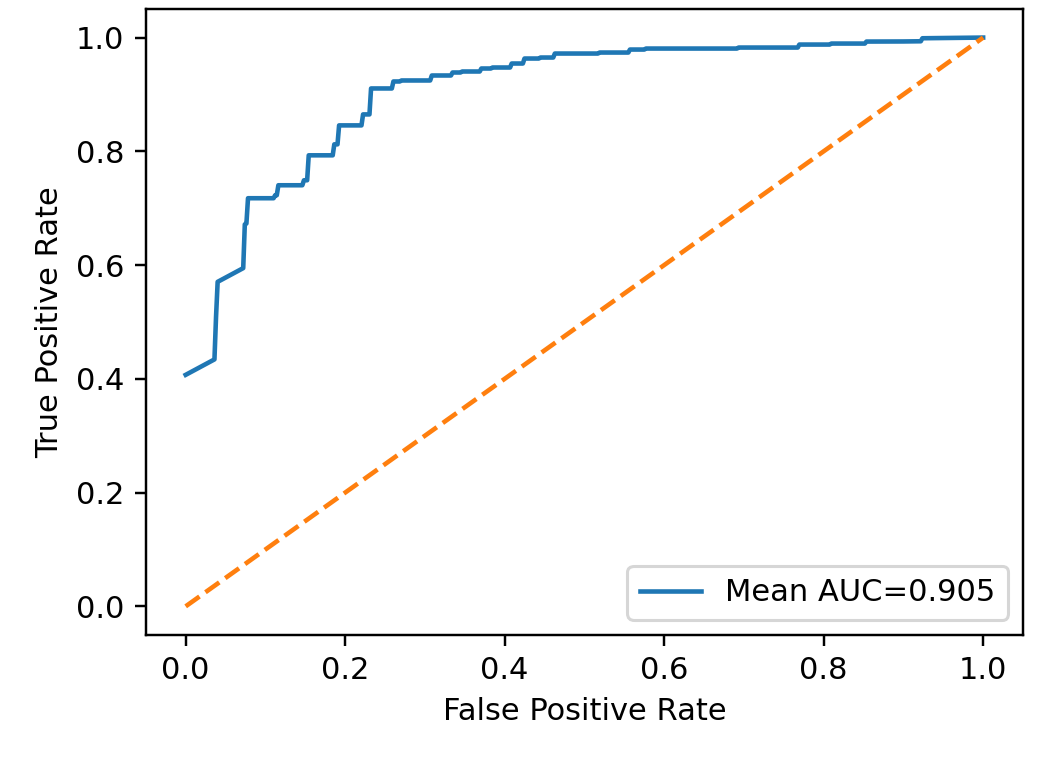}
  \caption{KNN: ROC curve (mean)}
  \label{fig:knn_roc}
\end{subfigure}

\caption{Performance visualisation of advanced ML classifiers (XGBoost, LightGBM, and KNN). For each model, the figure displays the averaged confusion matrix, mean precision–recall curve, 
and mean ROC curve, computed across 5-fold cross-validation.}

\label{fig:ml_group2}
\end{figure*}

\begin{figure*}
\centering
\begin{subfigure}[t]{0.32\textwidth}
  \centering
  \includegraphics[width=\linewidth]{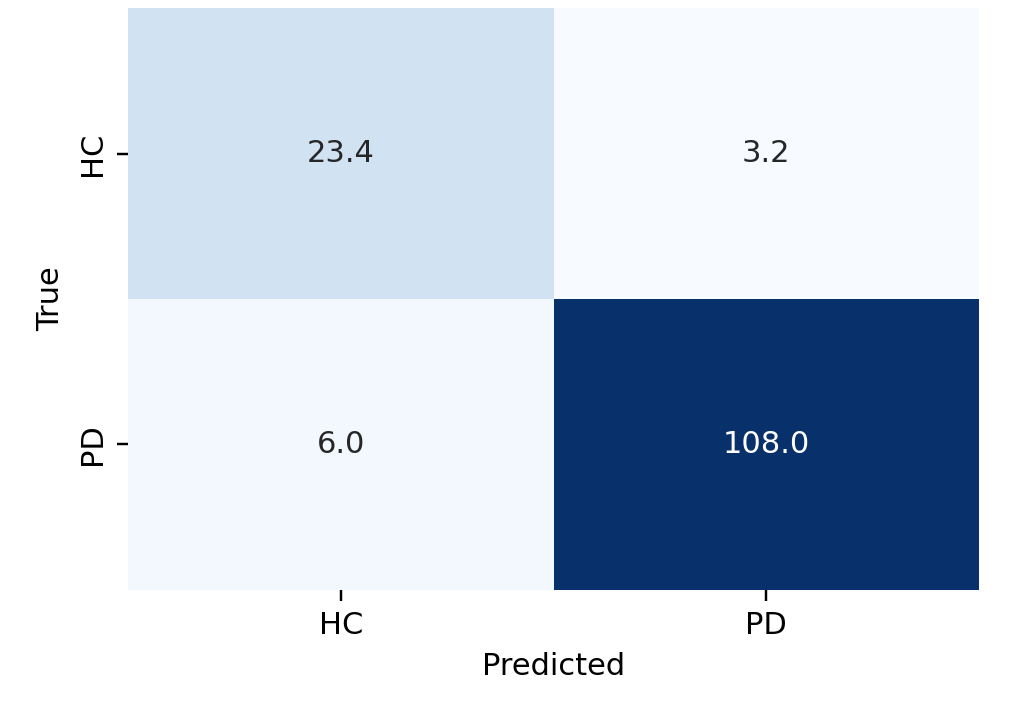}
  \caption{ANN: Confusion matrix (avg)}
  \label{fig:ann_confmat}
\end{subfigure}
\begin{subfigure}[t]{0.32\textwidth}
  \centering
  \includegraphics[width=\linewidth]{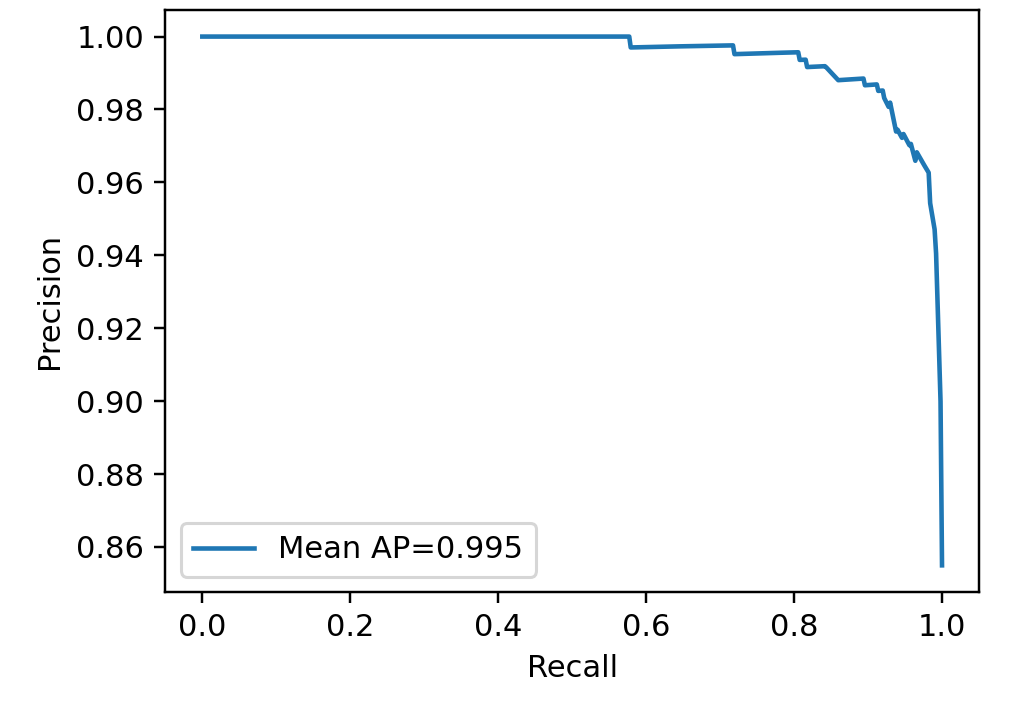}
  \caption{ANN: PR curve (mean)}
  \label{fig:ann_pr}
\end{subfigure}
\begin{subfigure}[t]{0.32\textwidth}
  \centering
  \includegraphics[width=\linewidth]{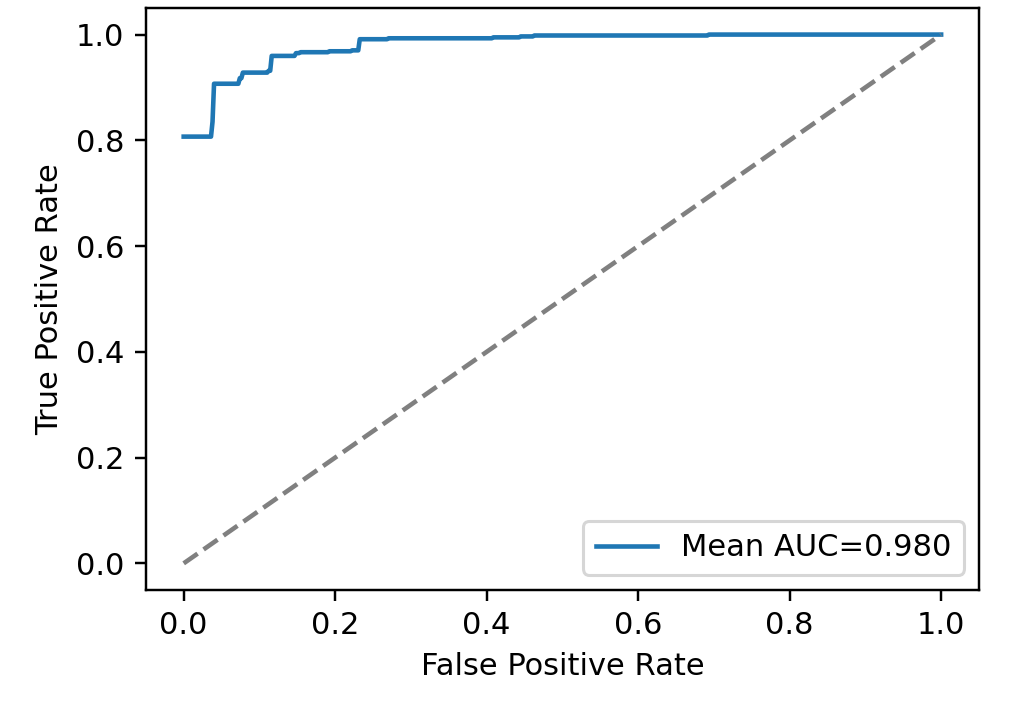}
  \caption{ANN: ROC curve (mean)}
  \label{fig:ann_roc}
\end{subfigure}

\vspace{6pt} 

\begin{subfigure}[t]{0.32\textwidth}
  \centering
  \includegraphics[width=\linewidth]{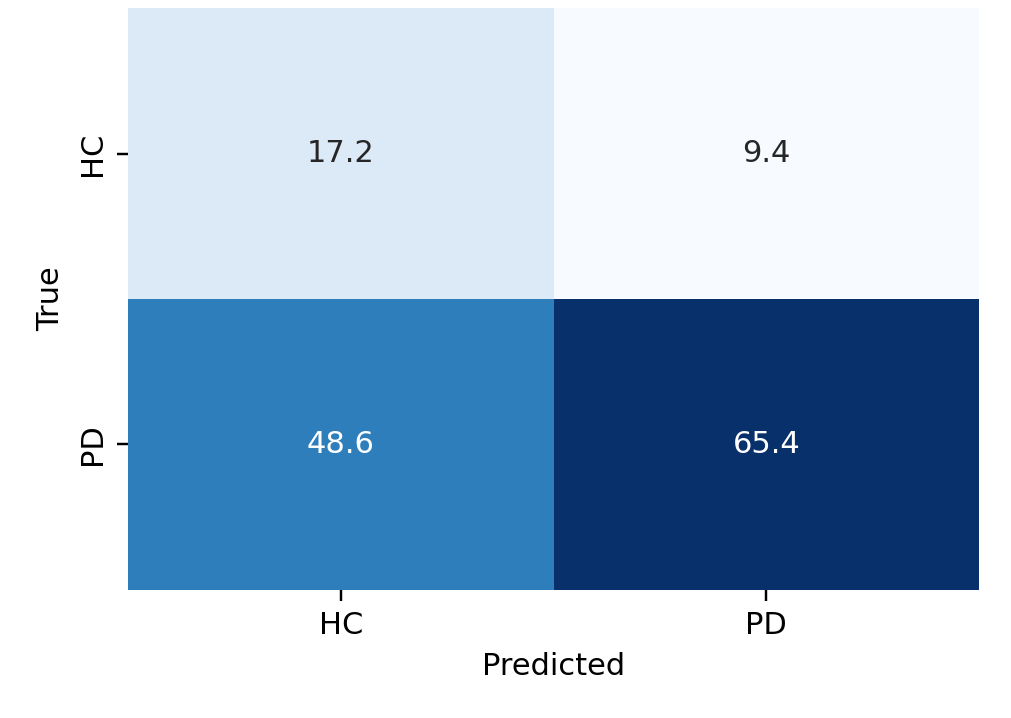}
  \caption{1d\textendash CNN: Confusion matrix (avg)}
  \label{fig:cnn1d_confmat}
\end{subfigure}
\begin{subfigure}[t]{0.32\textwidth}
  \centering
  \includegraphics[width=\linewidth]{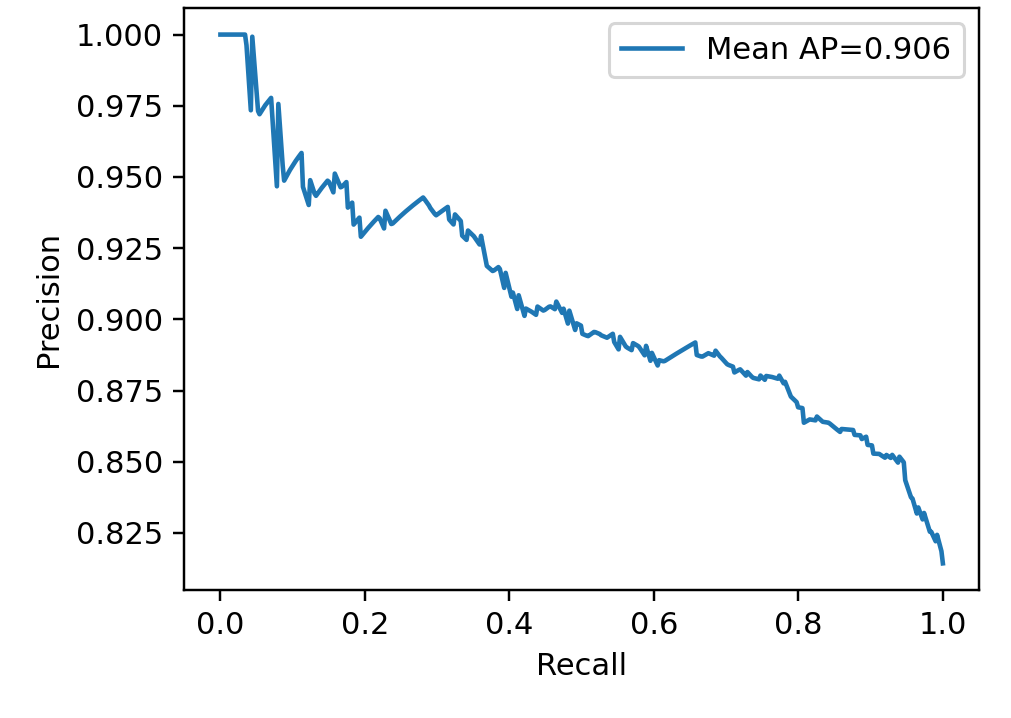}
  \caption{1d\textendash CNN: PR curve (mean)}
  \label{fig:cnn1d_pr}
\end{subfigure}
\begin{subfigure}[t]{0.32\textwidth}
  \centering
  \includegraphics[width=\linewidth]{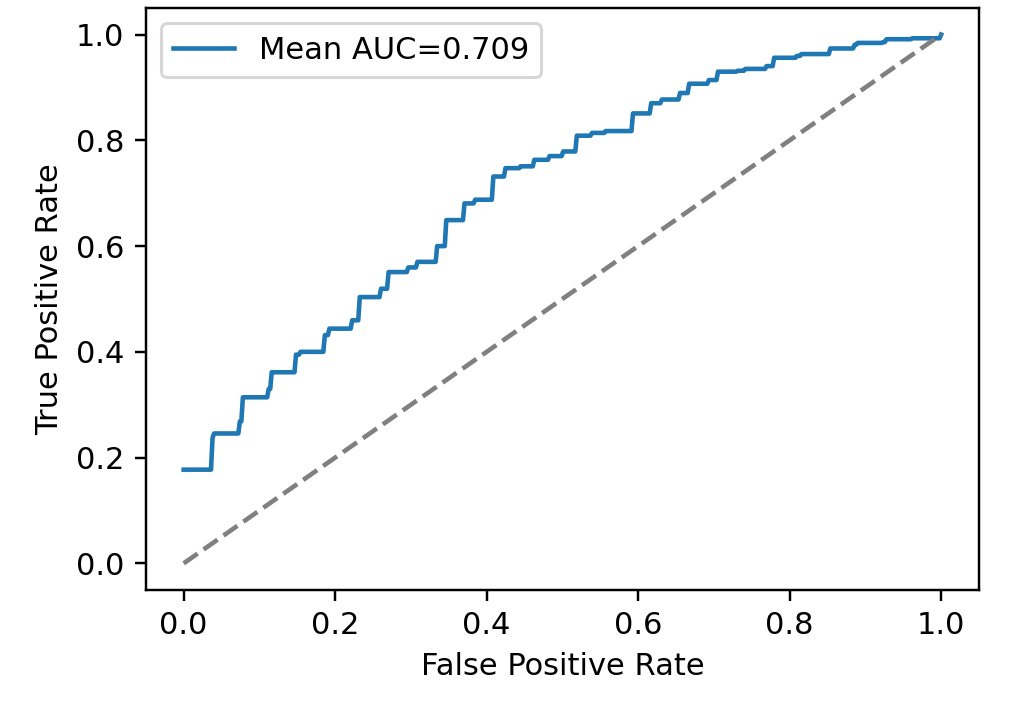}
  \caption{1d\textendash CNN: ROC curve (mean)}
  \label{fig:cnn1d_roc}
\end{subfigure}

\vspace{6pt}

\begin{subfigure}[t]{0.32\textwidth}
  \centering
  \includegraphics[width=\linewidth]{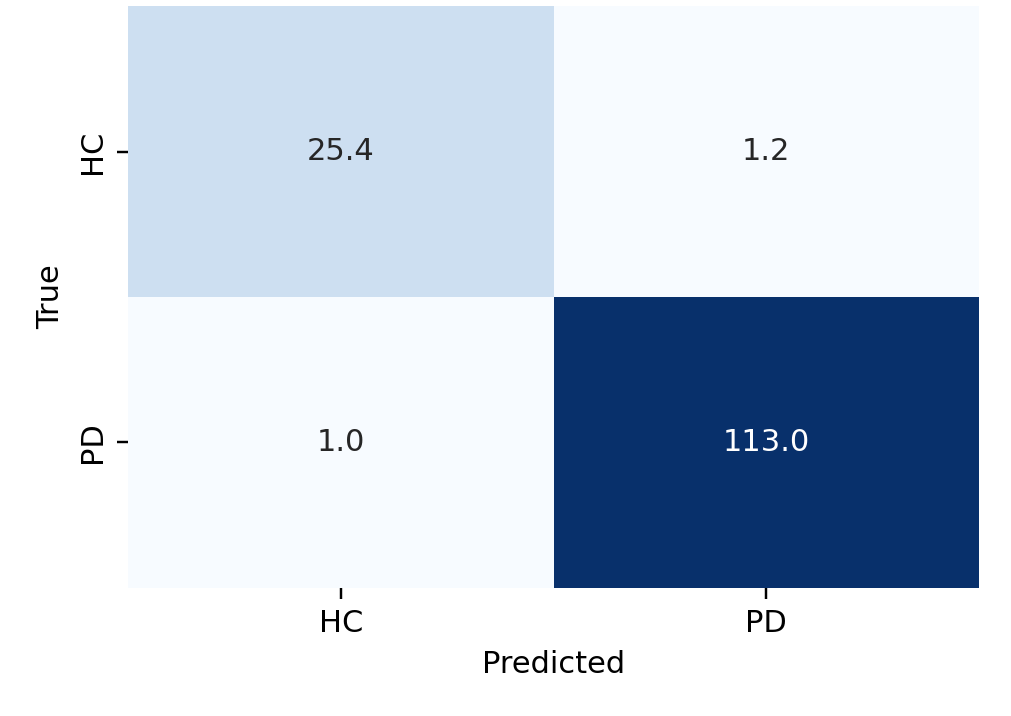}
  \caption{SAFN: Confusion matrix (avg)}
  \label{fig:safn_confmat}
\end{subfigure}
\begin{subfigure}[t]{0.32\textwidth}
  \centering
  \includegraphics[width=\linewidth]{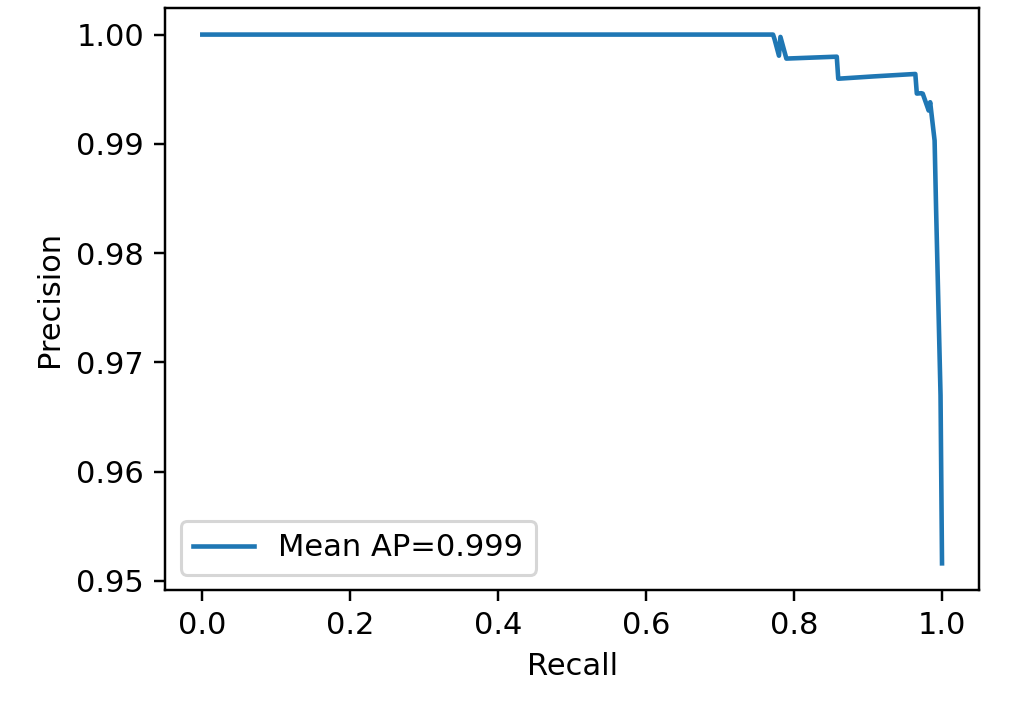}
  \caption{SAFN: PR curve (mean)}
  \label{fig:safn_pr}
\end{subfigure}
\begin{subfigure}[t]{0.32\textwidth}
  \centering
  \includegraphics[width=\linewidth]{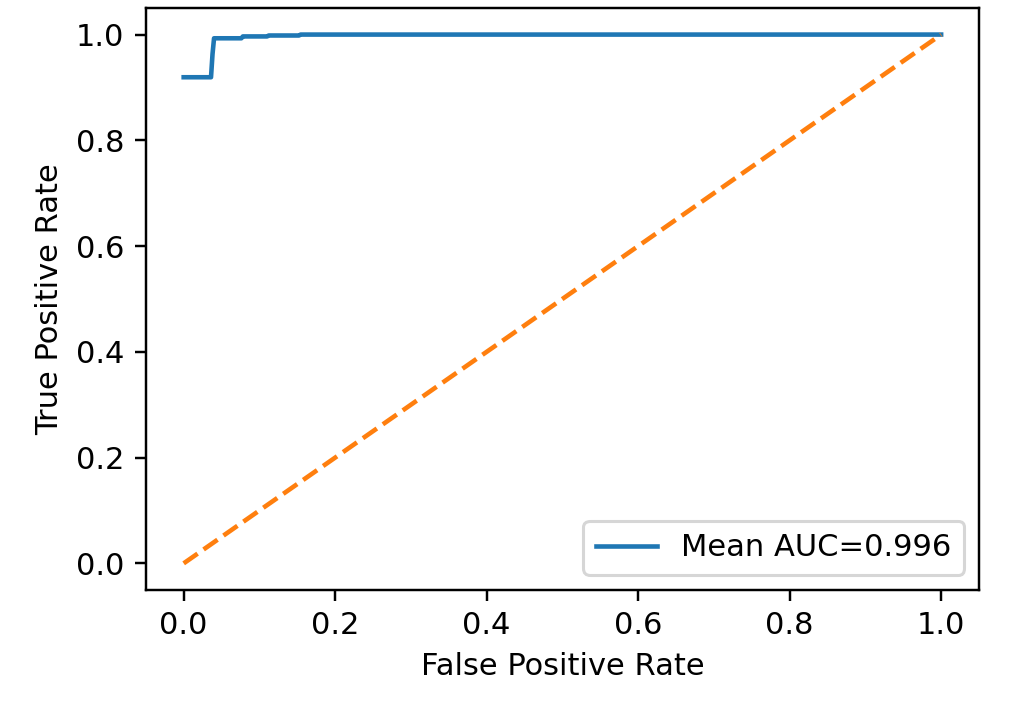}
  \caption{SAFN: ROC curve (mean)}
  \label{fig:safn_roc}
\end{subfigure}

\caption{Performance visualisation of deep learning models (ANN, 1D--CNN, and the proposed Class-Weighted SAFN). For each model, the figure displays the averaged confusion matrix, mean precision–recall curve, and mean ROC curve, computed across 5-fold cross-validation.}

\label{fig:dl_group}
\end{figure*}

This section presents the comparative results of all implemented models for PD classification. 
Both traditional ML and DL approaches were evaluated under a consistent 5-fold cross-validation framework to ensure fairness and reproducibility. 
Each model was assessed using complementary performance metrics—accuracy, balanced accuracy, ROC–AUC, PR–AUC, precision, recall, and F1-score—as described in the Methodology section.  
To maintain a fair comparison, all models were trained on the same feature set and identical data splits.  
Class imbalance was addressed using model-appropriate strategies: linear and kernel-based classifiers (Logistic Regression, SVM, KNN) were trained with \texttt{class\_weight=balanced}; tree-based ensembles (Random Forest, XGBoost, LightGBM) relied on built-in sample-reweighting; and DL models (ANN, 1D--CNN, SAFN) employed a class-weighted binary cross-entropy loss to ensure balanced learning between PD and HC samples.

Table~\ref{tab:model_performance} summarises the quantitative outcomes, followed by the grouped bar chart in Figure~\ref{fig:grouped_bar_performance} for a comparative visualisation of key metrics.  
The overall performance profiles across all seven metrics are further illustrated using the radar plot in Figure~\ref{fig:radar_performance}.  
Finally, Figures~\ref{fig:ml_group1}--\ref{fig:dl_group} provide detailed confusion matrices, PR curves, and ROC curves for each model group.

Across the ML models, most classifiers achieved high accuracy ($>$0.9), highlighting the discriminative strength of the multimodal feature representation derived from clinical, motor, and non-motor attributes.  
Among linear and kernel-based approaches, Logistic Regression and SVM produced nearly identical performance (accuracy: 0.94~$\pm$~0.01), each achieving ROC--AUC values above 0.97.  
Their stable performance reflects effective regularisation (via $L_2$ penalties and margin maximisation) and suitability for moderately nonlinear boundaries.  
Random Forest achieved slightly higher balanced accuracy (0.95~$\pm$~0.03) and F1-score (0.97~$\pm$~0.04), benefiting from its ensemble structure, which captures feature interactions and mitigates overfitting.  
However, its higher variance across folds (SD up to 0.06) likely results from sampling variability on modest fold sizes.

Among boosted tree-based models, XGBoost and LightGBM demonstrated the strongest and most consistent performance.  
Both models effectively captured nonlinear structure in the multimodal input space, but LightGBM achieved the highest accuracy (0.96–0.97) and balanced accuracy (0.96~$\pm$~0.01), attributable to its leaf-wise growth strategy and efficient regularisation.  
In contrast, KNN showed noticeably lower accuracy (0.88~$\pm$~0.01) and ROC--AUC (0.90~$\pm$~0.04), reflecting its sensitivity to high-dimensional feature scaling and the reduced discriminative value of distance-based metrics in heterogeneous feature spaces.

Figures~\ref{fig:ml_group1} and~\ref{fig:ml_group2} provide visual interpretation of these trends.  
Figure~\ref{fig:ml_group1} shows confusion matrices and PR/ROC curves for Logistic Regression, SVM, and Random Forest, all of which exhibit high true-positive rates and sharply rising PR curves, indicating near-perfect PD discrimination.  
Figure~\ref{fig:ml_group2} illustrates XGBoost, LightGBM, and KNN: the boosted models show compact confusion matrices and ROC curves approaching the top-left corner, while KNN presents wider PR curves consistent with its weaker quantitative performance.

Within the DL group (Figure~\ref{fig:dl_group}), the baseline ANN achieved strong results (accuracy: 0.93~$\pm$~0.03; ROC--AUC: 0.98~$\pm$~0.01), confirming that fully connected networks can model nonlinear dependencies effectively when appropriately regularised.  
The 1D--CNN, however, exhibited unstable performance (accuracy: 0.58~$\pm$~0.20), likely due to overfitting and limited benefit from convolutional structure in a feature space without strong local sequential dependencies.

The proposed Class-Weighted SAFN achieved the best overall performance across all models, with 0.98~$\pm$~0.02 accuracy and a perfect PR--AUC (1.00~$\pm$~0.00).  
SAFN integrates self-attention with feature normalisation, enabling dynamic reweighting of informative features and suppression of irrelevant or redundant dimensions.  
Its class-weighted optimisation further ensures balanced learning for minority and majority classes, resulting in improved recall and precision (both 0.99).  
The confusion matrix for SAFN is the most diagonally dominant among all models, demonstrating minimal misclassification and strong generalisation across folds.

Overall, the results in Table~\ref{tab:model_performance} and Figures~\ref{fig:ml_group1}--\ref{fig:dl_group} show that ensemble models such as LightGBM and XGBoost provide strong baselines for structured clinical data.  
Nevertheless, the proposed SAFN surpasses these baselines by leveraging attention-driven feature fusion and class-weighted optimisation, achieving superior generalisation and balanced prediction capability for PD classification.

\section{Discussion}
\label{sec:discussion}

\subsection{Statistical Group Differences}

Baseline group-level statistical analyses were performed to characterise differences between PD and HC participants prior to modelling. Continuous variables were first assessed for normality using the Shapiro--Wilk test. As most distributions deviated from normality, nonparametric Mann--Whitney~U tests were applied to compare PD and HC groups. Effect sizes were quantified using Cliff’s~$\delta$, where $|\delta|=0.2$, $0.5$, and $0.8$ represent small, medium, and large effects, respectively. Categorical variables were analysed using the $\chi^2$ test of independence, or Fisher’s exact test when any expected cell count was below~5. Associations for categorical variables were summarised using Cramér’s~$V$, interpreted as small~($V<0.1$), medium~($0.1\le V<0.3$), or large~($V\ge0.3$). All $p$-values were adjusted for multiple comparisons using the Benjamini--Hochberg false-discovery-rate (FDR) procedure with a significance threshold of $q=0.10$. Continuous variables are reported as median~[IQR], and categorical variables as counts and percentages.

\begin{table}[t]
\centering
\begin{threeparttable}
\caption{Top discriminative continuous variables at baseline (PD vs HC). Values shown as median [IQR]; $p_{\mathrm{FDR}}$ adjusted with Benjamini--Hochberg (q=0.10). Effect size is Cliff's $\delta$ (nonparametric; negative values indicate larger PD values for these variables).}
\begin{tabular}{lccccc}
\toprule
Variable & PD median [IQR] & HC median [IQR] & Test & $p_{\mathrm{FDR}}$ & Cliff's $\delta$ \\
\midrule
NP3TOT      & 22.00 [15.00, 30.00] & 1.00 [0.00, 3.00]  & Mann--Whitney U & 3.25e-54 & -0.98 \\
NHY         & 2.00 [1.00, 2.00]    & 0.00 [0.00, 0.00]  & Mann--Whitney U & 2.05e-77 & -0.94 \\
NP3BRADY    & 1.00 [1.00, 2.00]    & 0.00 [0.00, 0.00]  & Mann--Whitney U & 1.70e-47 & -0.84 \\
NP3FACXP    & 1.00 [1.00, 2.00]    & 0.00 [0.00, 0.00]  & Mann--Whitney U & 3.22e-46 & -0.79 \\
NP3RTCON    & 1.00 [0.00, 2.00]    & 0.00 [0.00, 0.00]  & Mann--Whitney U & 3.00e-34 & -0.66 \\
NP3FTAPR    & 1.00 [1.00, 2.00]    & 0.00 [0.00, 0.00]  & Mann--Whitney U & 1.59e-33 & -0.66 \\
NP3POSTR    & 0.00 [0.00, 1.00]    & 0.00 [0.00, 0.00]  & Mann--Whitney U & 2.79e-31 & -0.64 \\
NP3RIGRU    & 1.00 [0.00, 2.00]    & 0.00 [0.00, 0.00]  & Mann--Whitney U & 9.37e-31 & -0.61 \\
MSEADLG     & 90.00 [90.00, 100.00]& 100.00 [100.00, 100.00] & Mann--Whitney U & 2.93e-28 & 0.60 \\
NP3RIGLU    & 1.00 [0.00, 2.00]    & 0.00 [0.00, 0.00]  & Mann--Whitney U & 6.65e-28 & -0.60 \\
\bottomrule
\end{tabular}

\begin{tablenotes}[flushleft]
\footnotesize
\item \textit{Notes.} NP3TOT: total MDS-UPDRS Part III score; NHY: Hoehn \& Yahr stage; NP3BRADY: global bradykinesia; NP3FACXP: facial expression (hypomimia); NP3RTCON: rest tremor constancy; NP3FTAPR: finger tapping (right hand); NP3POSTR: posture; NP3RIGRU: rigidity (right upper limb); NP3RIGLU: rigidity (left upper limb); MSEADLG: Modified Schwab \& England Activities of Daily Living score (lower values indicate greater disability).
\end{tablenotes}

\label{tab:baseline_top_numeric}
\end{threeparttable}
\end{table}

\begin{table}[t]
\centering
\begin{threeparttable}
\caption{Categorical baseline differences (PD vs HC). $p_{\mathrm{FDR}}$ adjusted with Benjamini--Hochberg; effect size is Cramér's $V$.}
\renewcommand{\arraystretch}{1.25}
\begin{tabular}{p{5cm} p{10cm}}
\toprule
\textbf{Variable: PDSTATE} & \\
\midrule
\textbf{PD counts (\%)} 
    & OFF: 86 (15.1\%); ON: 173 (30.4\%); nan: 311 (54.6\%) \\[4pt]
\textbf{HC counts (\%)}  
    & OFF: 0 (0.0\%); ON: 0 (0.0\%); nan: 133 (100.0\%) \\[4pt]
\textbf{Test} & $\chi^2$ test \\[4pt]
\textbf{$p_{\mathrm{FDR}}$} & 4.22$\times$10$^{-20}$ \\[4pt]
\textbf{Cramér's $V$} & 0.37 \\[6pt]

\midrule
\textbf{Variable: COGCAT\_TEXT} & \\[-2pt]
\midrule
\textbf{PD counts (\%)} 
    & Cognitive Complaint: 20 (3.5\%); Dementia: 5 (0.9\%); MCI: 56 (9.8\%); Normal: 93 (16.3\%); nan: 381 (66.8\%) \\[4pt]
\textbf{HC counts (\%)} 
    & Cognitive Complaint: 2 (1.5\%); Dementia: 0 (0.0\%); MCI: 6 (4.5\%); Normal: 53 (39.8\%); nan: 72 (54.1\%) \\[4pt]
\textbf{Test} & Fisher/$\chi^2$ \\[4pt]
\textbf{$p_{\mathrm{FDR}}$} & 1.45$\times$10$^{-2}$ \\[4pt]
\textbf{Cramér's $V$} & 0.13 \\
\bottomrule
\end{tabular}

\begin{tablenotes}[flushleft]
\footnotesize
\item \textit{Notes.} PDSTATE: medication state at the time of the motor examination (OFF = withdrawal period; ON = after dopaminergic medication). COGCAT\_TEXT: cognitive diagnosis category (Normal, Mild Cognitive Impairment, Dementia, or self-reported Cognitive Complaint). “nan” indicates missing annotations in the PPMI baseline dataset.
\end{tablenotes}

\label{tab:baseline_categorical_rowwise}
\end{threeparttable}
\end{table}

Table~\ref{tab:baseline_top_numeric} summarises the most discriminative continuous variables showing statistically significant between-group differences after FDR correction. As expected, PD participants demonstrated markedly higher motor-severity scores, including UPDRS~III total (NP3TOT), Hoehn--Yahr stage (NHY), and bradykinesia (NP3BRADY), all with large to very large effect sizes (e.g.\ $|\delta|\ge0.84$) and extremely small adjusted $p$-values. Additional motor subscores such as NP3FACXP, NP3RTCON, NP3FTAPR, NP3POSTR, and NP3RIGRU also showed large effects. The non-motor functional measure MSEADLG exhibited a medium-to-large effect (Cliff’s~$\delta=0.60$), reflecting reduced independence in activities of daily living in the PD group.

These group differences match established and emerging evidence in PD research. NP3TOT, reflecting global motor severity, has repeatedly been identified as one of the strongest clinical predictors of PD progression \citep{germani2025predicting}. In a recent replication study, “Predicting Parkinson’s disease trajectory using clinical and functional MRI features,” MDS-UPDRS Part III metrics consistently formed the dominant predictive features, with model performance decreasing significantly when motor metrics were removed \citep{germani2025predicting}. This supports the prominence of NP3TOT as a high-information clinical marker and aligns directly with the large effect sizes observed in Table~\ref{tab:baseline_top_numeric}. Bradykinesia features (NP3BRADY, NP3FTAPR) are also highly relevant. A major multimodal explainable machine-learning study using five PPMI time-series modalities showed that bradykinesia was the most dominant and consistent predictor across multiple explainability frameworks (SHAP, LIME, SHAPASH) \citep{junaid2023explainable}. The authors concluded that bradykinesia-related scores were not only predictive but medically foundational for separating early PD from non-PD states, strongly supporting the bradykinesia-related effects observed here. The strong NHY separation likewise aligns with literature demonstrating that Hoehn--Yahr stage reflects progressive bilateral involvement, axial impairment, and postural instability, core diagnostic and prognostic indicators in PD \citep{rodriguez2013mds}. Similarly, the effect size for MSEADLG reflects functional decline, consistent with studies linking activities-of-daily-living measures to early indicators of reduced independence and increased disease burden \citep{lai2018clinical, shulman2006subjective}.

Facial expressivity (NP3FACXP) demonstrated a large group effect ($\delta=0.79$), consistent with hypomimia—reduced facial expression—as an early and sensitive motor sign of PD driven by dopaminergic and basal-ganglia dysfunction \citep{maycas2021hypomimia}. Explainable-AI studies have confirmed strong correlations between hypomimia, motor severity, and diagnostic utility \citep{filali2025explaining}. NP3RTCON, assessing \emph{rest tremor constancy}, showed a large effect size ($\delta=-0.66$), capturing a cardinal asymmetric symptom present in most newly diagnosed PD patients. Tremor reliably differentiates PD from healthy controls even when mild, aligning with patterns reported in early-stage drug-naive PD subtypes \citep{hou2018patterns} and recent characterisations of contralateral tremor spread \citep{pasquini2025contralateral}, as well as neurophysiological evidence of increased muscle tone \citep{asci2023rigidity,korkmaz2025upper}. Postural abnormalities assessed by NP3POSTR also showed a large effect ($\delta=-0.64$), reflecting axial impairment linked to disease severity, cognitive decline, and fall risk. Recent work using Kinect-derived postural indices has validated their discriminative power in separating PD from HC, mirroring our findings \citep{hong2022summary,pokhabov2024postural}.

The rigidity subscores NP3RIGRU and NP3RIGLU exhibited large effects ($\delta=-0.61$ and $\delta=-0.60$). Limb rigidity—typically asymmetric in early disease—is a cardinal motor feature of PD and appears in more than 80\% of de novo patients in PPMI. Rigidity consistently emerges as a discriminative MDS-UPDRS Part III item in Rasch-modelling studies and is strongly associated with phenotype classification and disease progression \citep{regnault2025optimizing,fereshtehnejad2017clinical}. Biomechanical and neurophysiological evidence has also linked upper-limb rigidity to overall motor disability \citep{asci2023rigidity,falletti2025rigidity}. Collectively, these findings reveal a coherent pattern of multidimensional motor impairment in PD, spanning rigidity, bradykinesia, axial symptoms, and functional independence. The large and consistent effect sizes across global motor (NP3TOT), limb, and fine-motor UPDRS subscores validate them as robust clinical markers of early PD and align with contemporary machine-learning studies confirming their foundational importance for PD identification and predictive modelling \citep{regnault2025optimizing}.

Table~\ref{tab:baseline_categorical_rowwise} presents categorical variables that differed significantly between PD and HC. PDSTATE exhibited a strong association with diagnosis (Cramér’s~$V=0.37$, $p_{\mathrm{FDR}}=4.22\times10^{-20}$), while the cognitive-category distribution (COGCAT\_TEXT) showed a smaller but significant association (Cramér’s~$V=0.13$, $p_{\mathrm{FDR}}=1.45\times10^{-2}$). These findings confirm that the dataset expresses expected clinical patterns of PD and is appropriate for subsequent predictive modelling. The clinical relevance of these categorical differences is also well supported. Medication state (ON/OFF) has well-established effects on measured motor performance and appears as a key contextual feature in multimodal modelling pipelines \citep{junaid2023explainable}, where medication history has been shown to enhance prediction accuracy. Cognitive impairment—spanning PD-MCI and PD dementia—is likewise well documented, with recent multimodal studies demonstrating that combining cognitive status with motor and functional measures substantially improves predictive modelling \citep{junaid2023explainable}. The same pattern is reflected here.

\subsection{Model-Derived Feature Importance from the Proposed SAFN}

To provide a fine-grained interpretation of the predictive behaviour of the proposed SAFN model, we conducted a feature-level attribution analysis using the Gradient~$\times$~Input method. 
Unlike model-agnostic perturbation approaches, Gradient~$\times$~Input directly quantifies the sensitivity of the model output to small changes in each input feature, via the element-wise product of the gradient and the input. 
This yields a differentiable estimate of how strongly each variable contributes to the predicted probability of PD. 
Attributions were computed on the validation split in each cross-validation fold, accumulated across all samples, and normalised so that feature-wise contributions sum to 100\%, which we report as percentage importance for interpretability.

\begin{figure}
    \centering
    \includegraphics[width=0.9\linewidth]{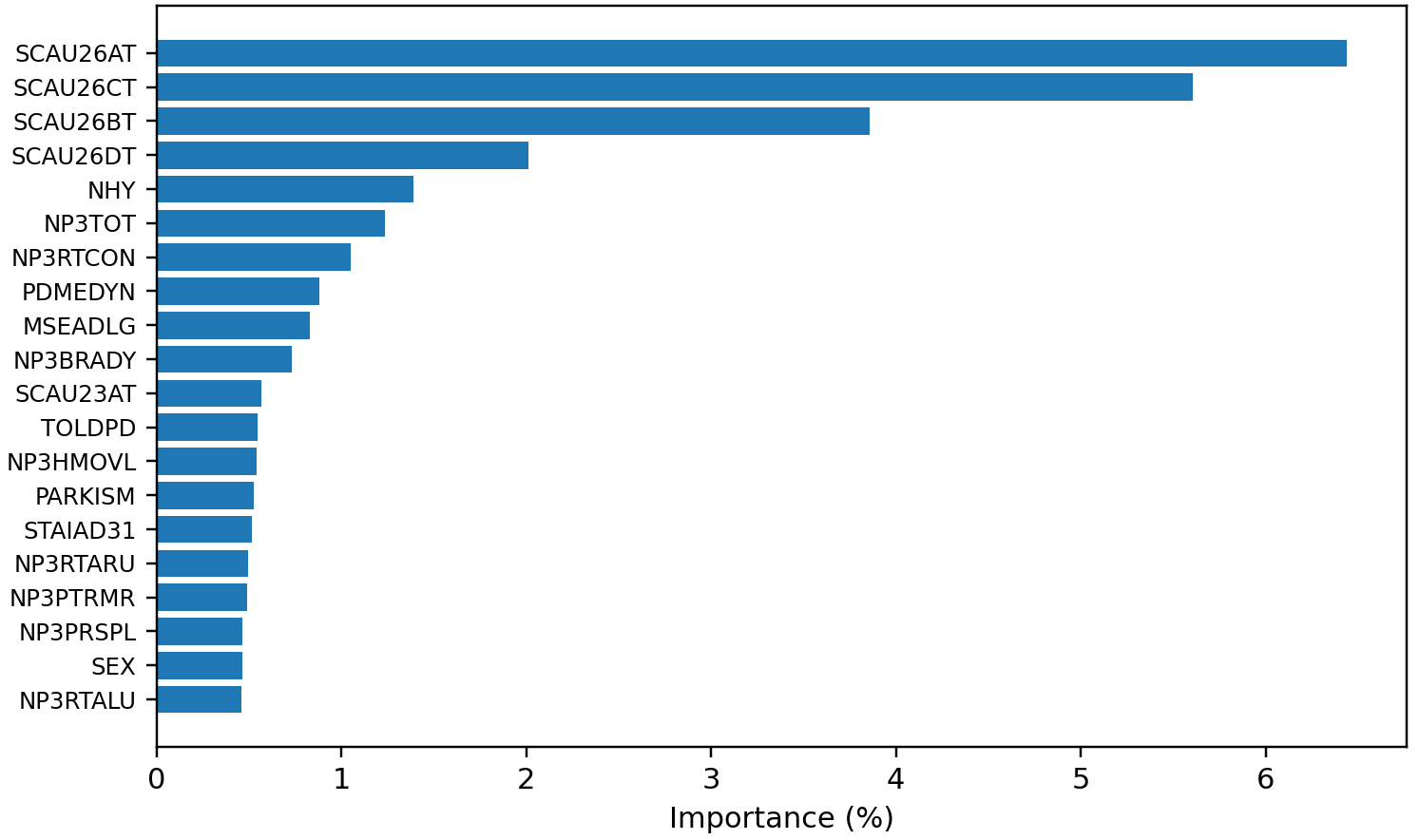}
    \caption{Top 20 most influential features identified by the proposed Class-Weighted SAFN model using the Gradient~$\times$~Input attribution method. Bars represent normalised feature contributions to the predicted PD probability, averaged across all folds and samples.}

    \label{fig:feat_importance_safn}
\end{figure}

Figure~\ref{fig:feat_importance_safn} presents the top 20 most influential features. 
A clear pattern emerges: clinical assessment variables overwhelmingly dominate the model’s decision-making, mirroring the modality-gate analysis where clinical features contributed approximately 60\% of the fused representation. 
Four autonomic symptom items (SCAU26AT, SCAU26CT, SCAU26BT, SCAU26DT) form the strongest predictors, jointly contributing more than 18\% of the total importance. 
These items correspond to key elements of the Scales for Outcomes in Parkinson’s Disease--Autonomic (SCOPA-AUT), consistent with the well-established role of autonomic dysfunction as an early and pervasive non-motor manifestation of PD.

Motor examination variables from the MDS-UPDRS Part~III also exhibit substantial importance. 
Notable contributions are observed for NP3TOT (overall motor severity), bradykinesia-related scores (NP3BRADY, NP3HMOVL), rigidity items (NP3RTCON, NP3RTARU, NP3RTALU), tremor scores (NP3PTRMR), and posture-related variables (NP3PRSPL). 
These findings reinforce the biological plausibility of SAFN, as motor impairment remains the central diagnostic hallmark of PD.

Global disease staging and diagnostic-status variables, including the Hoehn--Yahr stage (NHY), PD medication status (PDMEDYN), clinician-reported parkinsonism (PARKISM), and physician-confirmed diagnosis (TOLDPD), also register high importance. 
This pattern suggests that SAFN not only captures fine-grained motor and autonomic variations but also integrates broader disease context when discriminating between PD and healthy controls.

Interestingly, psychological and behavioural indicators (e.g., STAIAD31 from the State--Trait Anxiety Inventory) and demographic variables (SEX) also contribute measurably. 
These associations align with known epidemiological patterns: anxiety symptoms often co-occur with PD, and sex differences are well documented, with males showing a higher prevalence of PD.

Overall, the feature-importance analysis demonstrates that SAFN learns clinically grounded relationships: (i) autonomic dysfunction, (ii) motor impairment, and (iii) global disease status emerge as the most influential predictors, while demographic and behavioural variables provide secondary but complementary information. 
These results validate the interpretability and clinical relevance of the proposed model, offering a transparent view of its underlying decision mechanisms.

\subsection{Joint Interpretation: Statistical Effects vs.\ Learned Importances}

To provide a unified view of the PD–HC differences present in the dataset and how the proposed SAFN model leverages them, we compared the univariate statistical group differences (Tables~\ref{tab:baseline_top_numeric}--\ref{tab:baseline_categorical_rowwise}) with the model-derived feature attributions computed using Gradient~$\times$~Input (Figure~\ref{fig:feat_importance_safn}). Although both analyses examine PD–HC distinctions, they reflect fundamentally different perspectives: statistical tests quantify \emph{marginal} group differences for each feature independently, whereas SAFN’s attributions capture the \emph{multivariate, nonlinear} contributions learned within the full model.

A strong degree of agreement is observed between the two perspectives. Several variables with large univariate group differences—such as NP3TOT, NHY, NP3BRADY, NP3RTCON (rest tremor constancy), and MSEADLG—also appear among the highest-ranked SAFN features. These represent core motor and functional hallmarks of PD, including bradykinesia, rigidity, global motor severity, and daily-living impairment. Their consistency across both analyses reinforces the clinical validity of SAFN and shows that the model prioritises well-established PD symptoms.

However, SAFN also identifies patterns that do not emerge among the top univariate discriminators. Notably, four autonomic symptom items from the SCOPA–AUT scale (SCAU26AT, SCAU26CT, SCAU26BT, SCAU26DT) rank among the strongest predictors, despite only modest marginal group differences. This illustrates a key advantage of multivariate deep models: features that are individually weak but jointly informative become highly influential when interacting with related symptom domains. Autonomic dysfunction is widely recognised as an early and heterogeneous non-motor manifestation of PD; its prominence in SAFN suggests the model detects these subtle early-stage signals and integrates them with motor-domain markers.

Conversely, several features with strong univariate effects—such as NP3FACXP, NP3FTAPR (finger tapping, right hand), NP3POSTR, and other motor subscores—do not appear in SAFN’s top 20. This does not indicate a lack of utility; rather, SAFN likely down-weights these features due to redundancy with already-selected correlated motor variables that encode similar impairment patterns. Through tokenisation, self-attention, and gated fusion, SAFN compresses correlated signals and prioritises the most representative set of features for each symptom cluster.

Finally, variables such as PDMEDYN, TOLDPD, PARKISM, STAIAD31, and SEX also emerge as meaningful contributors in the attribution analysis. These show weaker univariate differences but gain relevance through interactions with motor and autonomic symptoms. Their inclusion aligns with known epidemiological and behavioural associations in PD, including sex differences in prevalence, anxiety-related traits, medication status, and clinician-confirmed diagnostic indicators.

Taken together, the comparison demonstrates that the statistical and model-based analyses are highly complementary. While univariate tests highlight the strongest \emph{individual} PD–HC differences, SAFN captures a richer \emph{interaction-driven} representation that integrates redundant motor patterns, subtle autonomic dysfunction, and contextual clinical information. The convergence between the two perspectives strengthens confidence in the interpretability and clinical plausibility of the proposed SAFN model.

\subsection{Modality Contribution Analysis}

To characterise how the proposed SAFN integrates heterogeneous information, we analysed the modality-level gate coefficients produced by the sparse attention–gated fusion layer. Each modality is assigned an independent sigmoid activation rather than a softmax-normalised weight, meaning that the four gate values do not necessarily sum to one. For interpretability, the raw gate coefficients from each fold were therefore normalised by dividing each value by the sum of all four gates. Table~\ref{tab:gates} reports both the averaged raw gate weights and their corresponding normalised percentage contributions.

\begin{table}[ht!]
\centering
\caption{Average modality-level gate weights from the proposed SAFN. Gate values are independent sigmoid-based scaling factors. Percentages represent normalised contributions across the four modalities.}
\label{tab:gates}
\small
\begin{tabular}{lcc}
\toprule
\textbf{Modality} & \textbf{Mean Gate Weight} & \textbf{Contribution (\%)} \\
\midrule
Clinical                & 0.118 & 59.9 \\
MRI Cortical Thickness  & 0.032 & 16.2 \\
Demographic             & 0.029 & 14.7 \\
MRI Volumetric          & 0.018 & 9.1  \\
\bottomrule
\end{tabular}
\end{table}

The results in Table~\ref{tab:gates} demonstrate that the clinical modality receives the highest relative weighting (approximately 60\%). This indicates that symptom severity scores, motor and non-motor assessments, functional measures, and medication-related variables provide the strongest discriminative information for distinguishing PD from HC in the PPMI cohort. This pattern is consistent with clinical knowledge: MDS-UPDRS scores, treatment status, and daily-living impairments directly reflect the core manifestations of Parkinsonism.

MRI Cortical Thickness and Demographic modalities contribute moderate weighting (16\% and 15\%, respectively). While cortical thickness differences between PD and HC are typically subtle at baseline, they provide complementary anatomical cues when combined with rich clinical symptomatology. Similarly, demographic factors such as age, sex, and handedness offer contextual information known to influence PD risk and phenotype expression.

The MRI Volumetric modality receives the lowest contribution (approximately 9\%). This suggests that global and regional volumetric measures—intracranial volume, grey/white matter volumes, and cerebellar metrics—carry relatively weak standalone discriminative signal in early PD. This observation is well supported by prior neuroimaging literature showing that volumetric atrophy patterns in PD are heterogeneous and more strongly associated with disease progression than with early case–control separation.

Overall, the gating analysis shows that SAFN \emph{adaptively prioritises modalities according to their learned predictive utility}. Clinical assessments dominate the decision process, while MRI (Cortical Thickness), demographic variables, and MRI (Volumetric) contribute complementary—but secondary—information. This behaviour enhances both predictive performance and interpretability by revealing how different modalities influence the fused representation used for classification.

\subsection{Ablation Study and Architectural Significance}

To evaluate the contribution of each modality and architectural component of the proposed SAFN model, a comprehensive ablation study was conducted using 5-fold cross-validation. The ablations included: (i) removing or isolating specific input modalities (MRI Cortical Thickness, Clinical, MRI Volumetric, and Demographic), (ii) removing individual architectural components such as cross-attention, modality gating, and class-weighting, and (iii) comparing against a plain MLP baseline trained on concatenated features. The quantitative results are summarised in Table~\ref{tab:ablation_safn}.

\begin{table}[t]
\centering
\caption{Comprehensive ablation study of the SAFN model under 5-fold cross-validation. The table evaluates the impact of selectively removing or isolating individual input modalities (MRI Cortical Thickness, Clinical, MRI Volumetric, and Demographic) as well as ablating key architectural components (cross-attention, modality gating, and class-weighting). A plain MLP trained on concatenated features is included as a baseline. All results are reported as mean~$\pm$~SD.}
\label{tab:ablation_safn}
\small
\setlength{\tabcolsep}{4pt}

\begin{tabular}{p{3.2cm} p{1.6cm} p{1.8cm} p{1.6cm} p{1.6cm} p{1.6cm} p{1.6cm} p{1.6cm}}
\toprule
\textbf{Model} & \textbf{Accuracy} & \textbf{Balanced Acc} & \textbf{ROC-AUC} & \textbf{PR-AUC} & \textbf{Precision} & \textbf{Recall} & \textbf{F1} \\
\midrule

Plain MLP (concat all features) 
& 0.94 $\pm$ 0.02 & 0.92 $\pm$ 0.03 & 0.97 $\pm$ 0.03 & 0.99 $\pm$ 0.01 & 0.97 $\pm$ 0.01 & 0.95 $\pm$ 0.02 & 0.96 $\pm$ 0.01 \\
\midrule

Clinical-only SAFN 
& 0.97 $\pm$ 0.01 & 0.93 $\pm$ 0.04 & 0.99 $\pm$ 0.00 & 1.00 $\pm$ 0.00 & 0.97 $\pm$ 0.02 & 0.99 $\pm$ 0.01 & 0.98 $\pm$ 0.01 \\

MRI Cortical Thickness--only SAFN 
& 0.68 $\pm$ 0.03 & 0.58 $\pm$ 0.04 & 0.61 $\pm$ 0.05 & 0.87 $\pm$ 0.02 & 0.85 $\pm$ 0.02 & 0.75 $\pm$ 0.05 & 0.79 $\pm$ 0.02 \\

SAFN w/o Clinical modality 
& 0.73 $\pm$ 0.02 & 0.65 $\pm$ 0.03 & 0.76 $\pm$ 0.03 & 0.94 $\pm$ 0.01 & 0.88 $\pm$ 0.02 & 0.77 $\pm$ 0.04 & 0.82 $\pm$ 0.02 \\

SAFN w/o MRI Cortical Thickness 
& 0.97 $\pm$ 0.02 & 0.95 $\pm$ 0.05 & 1.00 $\pm$ 0.01 & 1.00 $\pm$ 0.00 & 0.98 $\pm$ 0.02 & 0.98 $\pm$ 0.01 & 0.98 $\pm$ 0.01 \\
\midrule

SAFN w/o cross-attention 
& 0.97 $\pm$ 0.01 & 0.95 $\pm$ 0.03 & 1.00 $\pm$ 0.00 & 1.00 $\pm$ 0.00 & 0.98 $\pm$ 0.01 & 0.98 $\pm$ 0.01 & 0.98 $\pm$ 0.01 \\

SAFN w/o modality gates 
& 0.97 $\pm$ 0.02 & 0.95 $\pm$ 0.04 & 1.00 $\pm$ 0.01 & 1.00 $\pm$ 0.00 & 0.98 $\pm$ 0.02 & 0.99 $\pm$ 0.01 & 0.98 $\pm$ 0.01 \\

SAFN (no class-weighting) 
& 0.96 $\pm$ 0.02 & 0.92 $\pm$ 0.04 & 0.99 $\pm$ 0.01 & 1.00 $\pm$ 0.00 & 0.97 $\pm$ 0.02 & 0.98 $\pm$ 0.01 & 0.97 $\pm$ 0.01 \\
\midrule

\textbf{Full SAFN (proposed)} 
& \textbf{0.98 $\pm$ 0.02} & \textbf{0.97 $\pm$ 0.05} & \textbf{0.98 $\pm$ 0.02} & \textbf{1.00 $\pm$ 0.00} & \textbf{0.99 $\pm$ 0.03} & \textbf{0.99 $\pm$ 0.01} & \textbf{0.99 $\pm$ 0.01} \\
\bottomrule

\end{tabular}
\end{table}

\paragraph{Baseline comparison.}
The plain MLP baseline achieved an accuracy of 0.94~$\pm$~0.02 and ROC-AUC of 0.97~$\pm$~0.03, demonstrating that simple concatenation provides strong but clearly inferior performance compared with SAFN. This highlights the importance of SAFN’s architectural components—including feature tokenisation, attention-based encoding, and modality-aware fusion—which together capture richer feature interactions than a standard fully connected network.

\paragraph{Effect of modality subsets.}
A clear modality-dependent pattern emerged. The \textit{Clinical-only SAFN} achieved performance close to the full model (accuracy 0.97~$\pm$~0.01; F1-score 0.98~$\pm$~0.01), reflecting the high discriminative value of clinical assessments such as MDS-UPDRS, PDSTATE, medication status, and motor/non-motor symptoms.  
Conversely, the \textit{Cortical Thickness--only SAFN} performed substantially worse (ROC-AUC 0.61~$\pm$~0.05), consistent with prior findings that structural MRI differences between PD and HC are subtle at baseline and rarely sufficient for standalone classification.

Removing all clinical variables (\textit{SAFN w/o Clinical modality}) led to a marked drop in balanced accuracy (0.65~$\pm$~0.03), indicating that MRI Cortical Thickness, MRI Volumetric, and Demographic features provide complementary but comparatively weaker discriminative signals. In contrast, removing MRI Cortical Thickness (\textit{SAFN w/o MRI Cortical Thickness}) caused almost no degradation relative to the full model, confirming that clinical and demographic information dominate PD vs.\ HC separation in this cohort.

\paragraph{Effect of architectural components.}
Omitting cross-attention or modality gating resulted in small but consistent reductions in balanced accuracy and F1-score. Although the encoder captures strong intra-modality structure, cross-modality interactions (via cross-attention) and adaptive weighting (via gating) provide measurable performance gains. Removing class-weighting degraded balanced accuracy from 0.97 to 0.92, demonstrating the importance of class-balanced optimisation for handling the skewed PD/HC ratio.

\paragraph{Full model performance.}
The full SAFN achieved the strongest and most stable performance across all metrics: accuracy 0.98~$\pm$~0.02, balanced accuracy 0.97~$\pm$~0.05, and F1-score 0.99~$\pm$~0.01. Although several ablations reach similar ROC-AUC and PR-AUC values, these metrics saturate near ceiling. The decisive improvements appear in balanced accuracy and F1-score, confirming that the complete architectural design—tokenisation, attention encoding, cross-attention, sparse gating, and class-weighting—collectively yields the most robust multimodal classifier.

\paragraph{Interpretation.}
Overall, the ablation study reveals three key insights:  
(1) the Clinical modality provides the dominant discriminative signal;  
(2) MRI Cortical Thickness and MRI Volumetric features contribute limited standalone value but complement clinical information; and  
(3) SAFN’s architectural components (cross-attention, modality gating, class-weighted optimisation) enhance robustness and performance, even when improvements are incremental over strong clinical-only baselines.

These findings validate the design choices of SAFN and demonstrate its suitability for modelling heterogeneous multimodal biomedical data in PD classification.

\subsection{Comparison with Existing Literature}

A wide range of ML and DL approaches have been applied to PD classification across clinical, imaging, genetic, and multimodal datasets. Table~\ref{tab:literature_comparison_compact} summarises nine representative and recent studies that are directly relevant to the present work. These studies collectively highlight three major trends in the field: (i) strong predictive value of clinical and motor scales, (ii) increasing interest in multimodal fusion, and (iii) growing emphasis on interpretability, particularly for models that incorporate high-dimensional MRI data.

\begin{table*}[ht!]
\centering
\caption{Overview of recent PD classification studies using clinical, MRI, genetic, and multimodal data. The table summarises key modalities, model types, and performance metrics to contextualise the proposed SAFN against state-of-the-art approaches.}
\label{tab:literature_comparison_compact}
\small
\setlength{\tabcolsep}{3.5pt}

\begin{tabularx}{\textwidth}{p{3.5cm} p{2.9cm} p{3.5cm} p{1.7cm} p{3.2cm}}
\toprule
\textbf{Study} & \textbf{Data Modality} & \textbf{Model / Task} & \textbf{Performance} & \textbf{Notes} \\
\midrule

\citep{Esan2025Association} &
Clinical tabular (demographics, motor, cognitive, lifestyle) &
SVM, KNN, LR, RF, XGB, stacked ensemble; PD vs HC &
Acc = 0.93, AUC = 0.97 &
UPDRS, MoCA and functional scales strongest predictors; SMOTE used \\

\citep{Li2025PIDGN} &
Structural MRI + SNP genetics &
PIDGN (Transformer + 3D ResNet + gated attention); PD vs HC &
Acc = 0.86, AUC = 0.90 &
SHAP highlighted HLA-DRA and related SNPs; midbrain via Grad-CAM \\

\citep{Dentamaro2024} &
3D T1 MRI + clinical (age, sex, UPDRS-III) &
DenseNet + Excitation Network; PD vs HC &
Acc = 0.97, AUC = 0.99 &
Integrated Gradients showed prefrontal and temporal regions as salient \\

\citep{Yang2025} &
MRI + genetic SNP features &
Adaptive Ensemble Stacking (AE\_Stacking); PD vs HC &
Bal. Acc = 0.95, F1 = 0.93 &
SNCA- and VPS52-related variants most predictive \\

\citep{Camacho2023} &
Large multi-centre T1 MRI &
3D SFCN (lightweight CNN); PD vs HC &
Acc = 0.79, AUC = 0.87 &
Saliency maps: frontotemporal cortex and deep grey nuclei \\

\citep{Camacho2024Exploiting} &
T1 + DTI (FA, MD, RD, AD) &
3D CNN (SFCN variant); PD vs HC &
Acc = 0.81, AUC = 0.89 &
DTI metrics more salient than morphometry in saliency analysis \\

\citep{Alrawis2025} &
MRI (hybrid local + global features) &
FCN-PD (U-Net + EfficientNet + attention); PD vs HC &
Acc = 0.97--0.96 across datasets &
Attention maps highlight substantia nigra and cortical thickness patterns \\

\citep{Hussain2025} &
MRI slices (T1) &
TransPD-Net (CNN + Swin Transformer); PD vs HC &
Acc = 0.96, AUC = 0.96 &
Performance drops to AUC $\approx$ 0.67 on an unseen dataset \\

\citep{Basaia2024} &
MRI + clinical data &
3D CNN with transfer learning; PD severity vs HC &
Acc = 0.64--0.74 (mild / moderate--severe vs HC) &
CAMs: brainstem, temporal lobe, basal ganglia \\

\midrule
\textbf{This work (SAFN)} &
Multimodal tabular: MRI (Cortical Thickness), MRI (Volumetric), clinical scales, demographics (PPMI) &
Class-weighted SAFN with cross-attention and gated multimodal fusion; PD vs HC (5-fold CV) &
Acc = 0.98, AUC = 0.98 &
Modality gates and Grad$\times$Input highlight clinical scales (UPDRS-III, SCAU items, Hoehn--Yahr) as dominant predictors, with MRI and demographics providing complementary cues \\

\bottomrule
\end{tabularx}
\end{table*}

Several studies based purely on clinical or tabular data have achieved strong performance. For instance, Esan et al.~\citep{Esan2025Association} reported AUCs up to 0.97 using ensemble learning on demographic, motor, cognitive, and lifestyle features, with UPDRS and MoCA scores emerging as dominant predictors. This aligns with our findings, where disease-severity scales (e.g., SCAU items, UPDRS-III components, and Hoehn--Yahr stage) were also identified as the most influential features by the proposed SAFN model.

More complex multimodal studies incorporating structural MRI and genomics have recently emerged. Li et al.~\citep{Li2025PIDGN} proposed PIDGN, a hybrid Transformer--ResNet model integrating SNPs and MRI, achieving AUC = 0.90, while Yang et al.~\citep{Yang2025} used adaptive ensemble stacking to fuse MRI morphology with genetic variants, reaching balanced accuracy of 0.95. These results demonstrate the promise of combining complementary modalities, although these approaches rely primarily on high-dimensional MRI or SNP features and often require heavy preprocessing pipelines.

DL models based solely on MRI have also been investigated extensively. Camacho et al.~\citep{Camacho2023} and Camacho et al.~\citep{Camacho2024Exploiting} applied variants of the lightweight SFCN network to multi-centre T1 and DTI data, with AUC values between 0.87 and 0.89. More specialised MRI architectures, such as the FCN-PD model of Alrawis et al.~\citep{Alrawis2025} and the Transformer-based TransPD-Net~\citep{Hussain2025}, reported higher accuracies (0.96--0.97) but exhibited reduced generalisability on external datasets, confirming known robustness challenges in MRI-based PD classification. Similarly, multimodal MRI–clinical fusion approaches such as Dentamaro et al.~\citep{Dentamaro2024} and Basaia et al.~\citep{Basaia2024} achieved improved PD–HC separability but relied on computationally heavy 3D CNNs and required extensive image processing.

Relative to these studies, the proposed SAFN contributes three key advances. First, instead of relying on voxelwise MRI or end-to-end radiomics, SAFN integrates compact \emph{tabular} MRI-derived morphometry with clinical, demographic, and MRI (Volumetric) summaries, enabling efficient multimodal fusion with minimal preprocessing. Second, the model introduces a gated fusion mechanism and cross-attention layers designed specifically to learn inter-modality interactions—capabilities that are absent in traditional MLPs and many earlier fusion frameworks. Third, SAFN provides explicit interpretability through both modality-level gating and feature-level Grad$\times$Input attributions, enabling the identification of clinically meaningful predictors. The resulting importance profiles highlighted overall motor severity and bradykinesia (e.g., NP3TOT, NP3BRADY), axial/postural features, and autonomic SCAU26 items as dominant contributors, which closely mirror the strongest statistical group differences observed in the baseline analyses. This agreement between classical statistics and model-derived attributions enhances confidence in the physiological and clinical relevance of the learned representations.

Across the reviewed studies, SAFN achieves competitive or superior classification performance (Accuracy = 0.98, AUC = 0.98) while offering explicit interpretability and requiring substantially lower computational overhead than large 3D CNNs or Transformer-based MRI models. In addition, SAFN demonstrates strong robustness across folds and consistent feature attributions, further distinguishing it from prior multimodal PD classifiers where model stability or external generalisation was a recurring limitation.

Overall, this comparative analysis shows that the proposed SAFN complements and advances existing work by combining high accuracy, computational efficiency, multimodal sensitivity, and transparent interpretability within a single unified architecture.

\subsection{Limitations and Future Work}

Although the proposed Class-Weighted SAFN demonstrated strong performance and interpretability in distinguishing PD from HC participants, several limitations should be acknowledged.

First, the analysis was conducted on a single, harmonised PPMI-derived cohort of moderate size ($N = 703$; 570 PD, 133 HC) with a substantially imbalanced class distribution. While a class-weighted loss (Class-Balanced Focal Loss) was employed to mitigate this skew, reliance on a single dataset may introduce cohort-specific biases and limit external generalisability. Future work should therefore prioritise validation across larger, independent, multi-centre cohorts and diverse scanner platforms, and explore cross-dataset and cross-site generalisation through domain-adaptive or federated learning strategies.

Second, the study focused on static, cross-sectional data, whereas PD progression is inherently temporal. Longitudinal modelling of repeated assessments could enable prediction of disease trajectories, conversion risk, and progression rates. Extending SAFN with temporal architectures and uncertainty quantification would support progression forecasting, differential diagnosis, and personalised risk stratification.

Third, some modalities contained missing or partially available features, which may have constrained the model’s ability to fully exploit cross-modal relationships. Although SAFN can operate with incomplete tabular inputs, future research should examine principled data imputation, probabilistic modelling, or uncertainty-aware fusion mechanisms to better manage missingness and heterogeneous data availability. Scanner harmonisation and site-specific variability will also need to be addressed, particularly for multi-centre deployment.

Fourth, the MRI-derived inputs were based on preprocessed regional morphometry and volumetric summaries rather than raw 3D images. While this design enabled efficient multimodal integration, it may reduce sensitivity to fine-grained structural and microstructural alterations. Future extensions could incorporate lightweight imaging backbones or vision transformers operating directly on raw MRI or PET volumes, as well as integrate richer modalities such as microstructural MRI, electrophysiology, genetics, or digital phenotyping to capture multi-scale biomarkers.

Finally, although SAFN provides global interpretability through modality-level gating and feature-level attributions (e.g., Gradient~$\times$~Input), further work is required to translate these insights into patient-level, clinically actionable explanations. Post-hoc techniques such as SHAP, saliency maps, or Grad-CAM, combined with user-centred interface design, may produce clearer, human-readable explanations suitable for clinical decision support. Prospective, user-centred evaluations and real-world benchmarking will be essential to assess workflow integration, clinical usability, and regulatory readiness.

Overall, future research should focus on (i) external and cross-site validation, (ii) longitudinal and temporal extensions, (iii) richer and more complete multimodal inputs including raw imaging, and (iv) patient-centred interpretability, with the overarching goal of advancing SAFN toward robust, real-world clinical decision support.

\section{Conclusion}
\label{sec:conclusion}

This study presented the Class-Weighted SAFN, an interpretable multimodal DL framework designed to address key limitations in computational PD diagnosis, including class imbalance, modality heterogeneity, and the risk of clinically meaningful information being overshadowed by high-dimensional MRI features. By combining structural MRI Cortical Thickness, MRI Volumetric, rich motor and non-motor clinical assessments, and demographic data within a unified attention-based architecture, SAFN achieves biologically coherent and context-aware multimodal integration.

Central to the framework is a sparsity-regularised, attention-gated fusion layer paired with a symmetric cross-attention mechanism, enabling selective and parsimonious weighting of heterogeneous inputs. These components promote balanced representation learning across modalities and enhance the interpretability of the fused model. The hierarchical attribution analysis further demonstrated that SAFN captures clinically meaningful patterns, with global modality weighting and fine-grained feature attributions aligning closely with established PD symptomatology and diagnostic principles.

Overall, the proposed SAFN framework illustrates the value of carefully structured attention-based multimodal fusion for PD classification, offering a robust, interpretable, and computationally efficient alternative to traditional high-dimensional imaging pipelines. Its design provides a strong foundation for future extensions to longitudinal modelling, multi-centre validation, and real-world decision-support applications in clinical neurology.

\section*{Acknowledgment}
This work has been financially supported by the AI and Cyber Futures Centre (AICF), whose activities are funded primarily by Charles Sturt University, Australia.

\section*{Declaration of Generative AI and AI-assisted Technologies in the Writing Process}

During the preparation of this work the author(s) used Grammarly in order to proofread the manuscript and improve the clarity of the English writing. After using this tool/service, the author(s) reviewed and edited the content as needed and take(s) full responsibility for the content of the published article.

\bibliographystyle{cas-model2-names}

\bibliography{cas-refs}

\end{document}